%% file: main.tex
\documentclass[conference]{IEEEtran}
\pdfoutput=1
\IEEEoverridecommandlockouts

\usepackage{amssymb, subfigure, setspace, multirow}
\usepackage{verbatim}
\usepackage{amsmath}    
\usepackage{hhline}     
\usepackage{cite}
\usepackage{algorithmic}
\usepackage{algorithm}
\usepackage{array}
\usepackage{url}
\usepackage{bm}
\usepackage{mathrsfs}
\usepackage{amsfonts}
\usepackage{graphicx}
\usepackage{textcomp}
\usepackage{xcolor}
\usepackage{microtype}
\usepackage{gensymb}
\usepackage{balance}
\usepackage{cleveref}
\usepackage{color}
\usepackage{graphicx,times}
\usepackage{epstopdf}
\usepackage{indentfirst}
\usepackage{CJK}
\usepackage{txfonts}
\usepackage{theorem}
\usepackage{fancyhdr}

\crefname{section}{\S}{\S\S}
\Crefname{section}{\S}{\S\S}

\newcommand{\ie}{{\em i.e.}}
\newcommand{\eg}{{\em e.g.}}
\newcommand{\et}{{\em et al.}}

\newcommand\inv[1]{#1\raisebox{1.15ex}{$\scriptscriptstyle-\!1$}}

\def\spth{\textsuperscript{th}}

\def\BibTeX{{\rm B\kern-.05em{\sc i\kern-.025em b}\kern-.08em
		T\kern-.1667em\lower.7ex\hbox{E}\kern-.125emX}}

\begin{document}
\title{RF Backscatter-based State Estimation for Micro Aerial Vehicles}



\author{\IEEEauthorblockN{Shengkai Zhang,  Wei Wang{${^\ast}$}, Ning Zhang, Tao Jiang}
	\ School of Electronics Information and Communications, and Wuhan National Laboratory for Optoelectronics
	\\ Huazhong University of Science and Technology, China
	\\ Email: \{szhangk, weiwangw, ning\_zhang, taojiang\}@hust.edu.cn
	\thanks{${^\ast}$The corresponding author is Wei Wang (weiwangw@hust.edu.cn).}
}

\maketitle
\pagestyle{fancy}
\thispagestyle{fancy}          
\fancyhead{}                     
\chead{IEEE INFOCOM 2020 - IEEE International Conference on Computer Communications}
\cfoot{} 
\renewcommand{\headrulewidth}{0pt}     
\renewcommand{\footrulewidth}{0pt}






\begin{abstract}
The advances in compact and agile micro aerial vehicles (MAVs) have shown great potential in replacing human for labor-intensive or dangerous indoor investigation, such as warehouse management and fire rescue. However, the design of a state estimation system that enables autonomous flight in such dim or smoky environments presents a conundrum: conventional GPS or computer vision based solutions only work in outdoors or well-lighted texture-rich environments. This paper takes the first step to overcome this hurdle by proposing Marvel, a lightweight RF backscatter-based state estimation system for MAVs in indoors. Marvel is nonintrusive to commercial MAVs by attaching backscatter tags to their landing gears without internal hardware modifications, and works in a plug-and-play fashion that does not require any infrastructure deployment, pre-trained signatures, or even without knowing the controller's location. The enabling techniques are a new backscatter-based pose sensing module and a novel backscatter-inertial super-accuracy state estimation algorithm. We demonstrate our design by programming a commercial-off-the-shelf MAV to autonomously fly in different trajectories. The results show that Marvel supports navigation within a range of $50$ m or through three concrete walls, with an accuracy of $34$ cm for localization and $4.99\degree$ for orientation estimation, outperforming commercial GPS-based approaches in outdoors.
\end{abstract}


\section{Introduction}
\label{sec:intro}
\input{intro}

\section{System Overview}
\label{sec:overview}
\input{overview}

\section{Backscatter-based Pose Sensing}
\label{sec:lr}
\input{chirp}

\section{Backscatter-inertial Super-accuracy State Estimation}
\label{sec:pose}
\input{state_estimation}

\section{Implementation and Evaluation}
\label{sec:evaluation}
\input{eval}

\section{Related Work}
\label{sec:related}
\input{related}

\section{Conclusion}
\label{sec:conclusion}
\input{conclusion}

\bibliographystyle{IEEEtran}
\bibliography{main}

\end{document}

%% file: intro.tex
Over the last decade, the rapid proliferation of micro aerial vehicles (MAV) technologies has shown great potential in replacing human for labor-intensive or even dangerous indoor investigation and search, such as warehouse inventory management and fire rescue~\cite{ma2017drone, mao2017indoor, yang2019imgsensingnet, lin2018autonomous, dhekne2019trackio}. Specifically, using MAVs to manage inventory for warehouses cuts inventory checks from one month down to a single day~\cite{walmart_drone}, and using MAVs for search and rescue in firefighting operations saves the lives of firefighters by the fact that $53\%$ of deaths of the firefighters in the United States occurred in burning buildings in 2017~\cite{firefighter}. These applications require MAVs navigating autonomously in dim warehouses~\cite{fichtinger2015assessing} or smoky buildings while reporting to a server or controller at a distance or through walls. 

{\em State estimation} is fundamental to the autonomous navigation of MAVs. The state, including position, velocity, and orientation, is the key to the flight control system of an aerial vehicle that adjusts the rotating speed of rotors to achieve desired actions for responding remote control or autonomous operations. The mainstream uses GPS, compass and vision sensors to estimate a MAV's state. However, GPS-compass based approaches~\cite{farrell2008aided, chao2010autopilots} only work in outdoor free space since GPS signals can be blocked by occlusions and compass measurements are easily distorted by surrounding environments. In indoors, computer vision (CV) based approaches have attracted much attentions due to their lightweight, high accuracy, and low cost, while limited to good lighting or texture-rich environments~\cite{zhu2017event, dong2019pair, qin2017vins, mur2015orb}, thereby failing to work in dim warehouses or smoky fire buildings. 

Recent years have witnessed much progress in using RF signals to track a target's pose (position and orientation), holding the potential to state estimation that is highly resilient to visual limitations. Despite novel systems that have led to high accuracy~\cite{liu2017cooperative, zhang2018wins, luo20193d, jiang20193d, wei2016gyro, yang2014tagoram, wu2019sigcomm_rim, shangguan2016design}, hardly any of these ideas have made it into the scenario of indoor MAVs that needs the following requirements:

\begin{itemize}
	\item {\bf Long range/through wall.} To scan items across a warehouse or navigate in a fire building, the system should support the navigation at least across rooms or over an area of tens of meters.
	\item {\bf Lightweight.} As a MAV is typically compact with limited battery capacity, it requires a lightweight, small-sized, and low-power sensing modality to enable state estimation. 
	\item {\bf Plug-and-play.} To make it practical to emergency rescue and efficient indoor investigation, the system should be instantly operational in an unknown environment without prior infrastructure deployment, pre-training process, or labor-intensive setup. 
\end{itemize}

\begin{figure}
	\centering
	\begin{minipage}[b]{0.49\textwidth}\centering
		\center
		\includegraphics[width=1\textwidth]{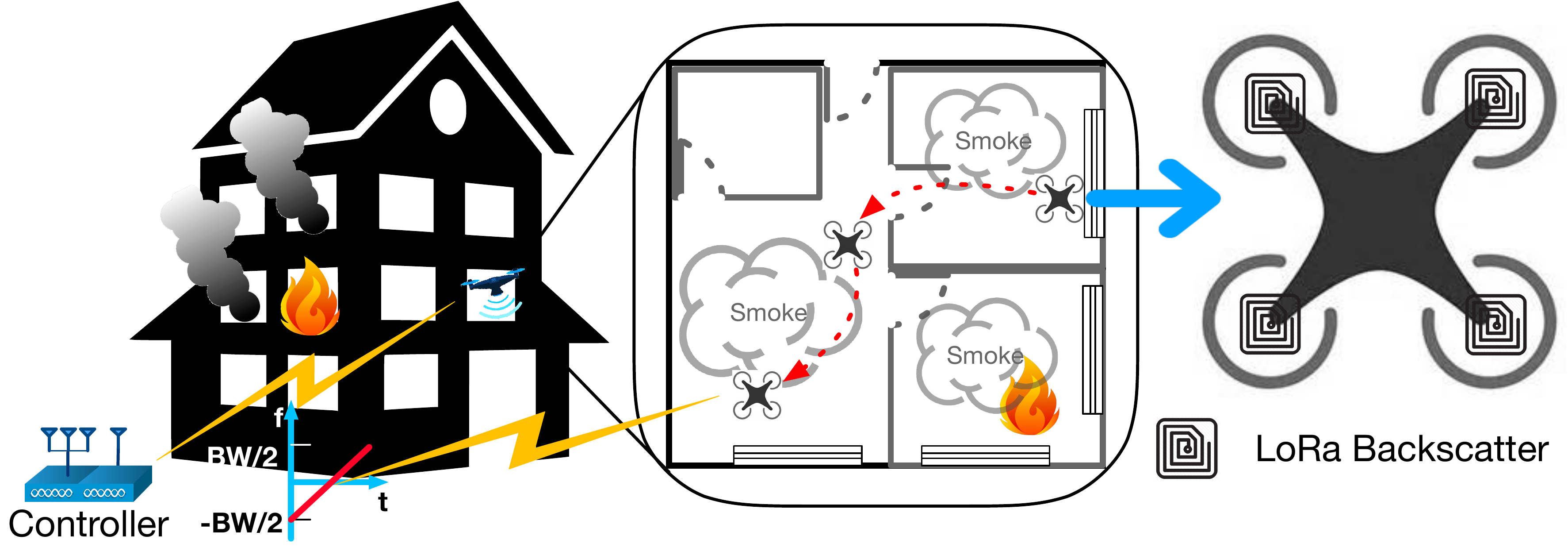}
		\caption{Usage example.} \label{fig:toy}
	\end{minipage}
\end{figure}
\noindent Specifically, RFID-based solutions~\cite{luo20193d, jiang20193d, wei2016gyro, yang2014tagoram, shangguan2016design} are lightweight while their operational range is limited. UWB and WiFi based solutions~\cite{liu2017cooperative, zhang2018wins, wu2019sigcomm_rim} have better operational range while requiring pre-deploying multiple anchors to enable the localizability, failing to meet the plug-and-play requirement. 

In this paper, we present Marvel, a state estimation system for MAVs that satisfies all these requirements. As shown in Fig.~\ref{fig:toy}, Marvel is able to support the navigation in burning buildings with smoke and fog. It allows a MAV to safely fly across rooms or in an area of several tens of meters. The system is lightweight in that it merely attaches a few backscatter tags to the landing gear of a commercial MAV without any internal hardware modification. Marvel works in a plug-and-play fashion that only requires the MAV's single controller, whose location is unknown, to interact with the MAV and the onboard backscatter tags. The design of Marvel is structured around two components: 
 
 {\bf (a) Backscatter-based pose sensing:} 
 Marvel's first component enables a backscatter-based sensing modality that allows it to estimate the response of the attached tags over backscattered signals that are drowned by noise. This sensing modality leverages chirp spread spectrum (CSS) signals to enable the pose tracking in long range or through occlusions where the signal amplitudes and phases are not available. It introduces a set of algorithms that first estimate channel phases of the tags' backscatter signals under mobility and then use the phases to estimate pose features, including the range, angle and rotation of the MAV to its controller. This component enables Marvel to operate across rooms or in an area of several tens of meters. 
 
{\bf (b) Backscatter-inertial super-accuracy state estimation:} The estimates of the first component cannot fulfil the state estimation, \ie, computing a MAV's position, velocity, and orientation. Marvel's second innovation is a backscatter-inertial super-accuracy algorithm that combines the backscatter-based estimates with the onboard IMU measurements to enable accurate state estimation. Despite that IMU suffers in error accumulation, our backscatter sensing is drift-free, \ie, no temporal error accumulation, being able to correct the IMU drift by multi-sensor fusion. It employs a graph-based optimization framework to compute a Gaussian approximation of the posterior over the MAV trajectory. This involves computing the mean of this Gaussian as the configuration of the state that maximizes the likelihood of the sensor observations.

{\bf Results}. We build a prototype of Marvel using a DJI M100 MAV attached with four LoRa backscatters customized by off-the-shelf components. We demonstrate Marvel's potential by programming the MAV to autonomously fly in different trajectories in a long-range open space and an across-room indoor test site. The results show that Marvel can support the navigation within a range of $50$ m or with three concrete walls blocked the vehicle to its controller and achieves an average accuracy of $34$ cm for localization and $4.99\degree$ for orientation estimation, outperforming commercial GPS-based navigation approaches~\cite{farrell2008aided, chao2010autopilots}.

{\bf Contributions}. We introduce a novel backscatter-based pose sensing technique that first extracts the channel phases of CSS signals under mobility and then estimates pose features of a MAV by the phases. This enables pose tracking in long range or through occlusions. In addition, we design a backscatter-inertial super-accuracy algorithm that fuses backscatter-based estimates and IMU measurements to enable accurate state estimation. Finally, we implement a prototype and conduct real-world experiments demonstrating the system's ability to navigate a MAV across rooms or in an area of several tens of meters.

%% file: overview.tex
\begin{figure}
	\centering
	\begin{minipage}[b]{0.49\textwidth}\centering
		\center
		\includegraphics[width=1\textwidth]{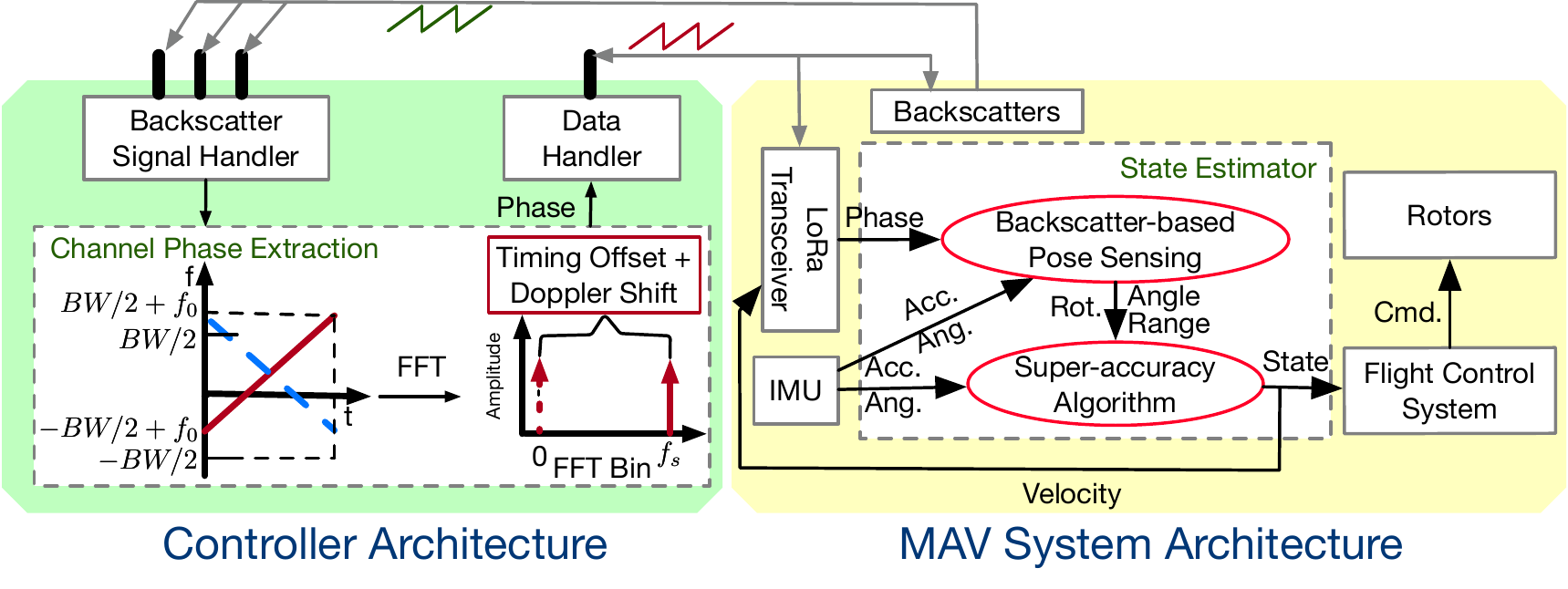}\vspace{-0.3cm}
		\caption{System architecture.} \label{fig:overview}
	\end{minipage}
\end{figure}

Marvel is a system that enables state estimation to support a MAV's navigation over an open area of tens of meters or across rooms in indoors. It works with the MAV's controller and four LoRa backscatter tags attached on the landing gear. As shown in Fig.~\ref{fig:overview}, the controller consists of a data handler and a backscatter signal handler. The data handler sends and receives data to and from the MAV by one antenna. It excites the tags and exchanges data like channel phases and velocities via CSS signals of LoRa. The backscatter signal handler receives the backscattered signals by a linear array of three antennas. It extracts the channel phases and then sends them to the MAV. The computer on the MAV takes the phases to compute the state. It then sends the state to the flight control system that computes commands to adjust actions and navigates the MAV to destination.

Marvel consists of two components: 
\begin{itemize}
	\item {\bf Backscatter-based pose sensing module} contains chirp-based estimation approaches that first estimate channel phases and then use the phases to compute the range, angle and rotation of the MAV to its controller, enabling pose tracking in long range or through occlusions.
	\item {\bf Backscatter-inertial super-accuracy state estimation algorithm} fuses the estimates of the backscatter-based sensing with the IMU measurements, which include 3D accelerations and angular velocities, through a graph-based optimization framework. It models each module's estimates as a Gaussian mixture and computes a Gaussian approximation of the posterior over the MAV trajectory. 
\end{itemize}

%% file: chirp.tex

\subsection{Primer on CSS Processing}
\label{subsec:primer}
Before proceeding to our design, we first briefly introduce the CSS signals of LoRa that we use. The controller transmits a linear upchirp signal with a bandwidth $BW$ to the MAV with backscatter tags. A tag backscatters the signal with a frequency shift $f_0$ for preventing the interference from the excitation signal. Multiple tags perform different frequency shifts. Marvel uses the chirp signal that is compatible with LoRa protocol. Its duration $T$ depends on spreading factor ($SF$) and bandwidth~\cite{liando2019known}, \ie, $T = 2^{SF}/BW$, where $SF \in \{6, 7, 8, 9, 10, 11, 12\}$ and the maximum $BW$ is $500$ KHz. 

To decode the chirp, the receiver first multiples the received signal with a synthesized downchirp whose frequency linearly varies from $BW/2 + f_0$ to $-BW/2 + f_0$. Then, it takes a fast Fourier transform (FFT) on this multiplication (Fig.~\ref{fig:overview}). This operation sums the energy across all the frequencies of the chirp, producing a {\em peak} at an FTT bin. 

Next, we describe how the controller extracts the channel phase of the received chirp from the peak. Then, we elaborate on how Marvel leverages the phase to estimate its position and rotation.

\begin{figure}
	\centering
	\begin{minipage}[b]{0.35\textwidth}\centering
		\center
		\includegraphics[width=1\textwidth]{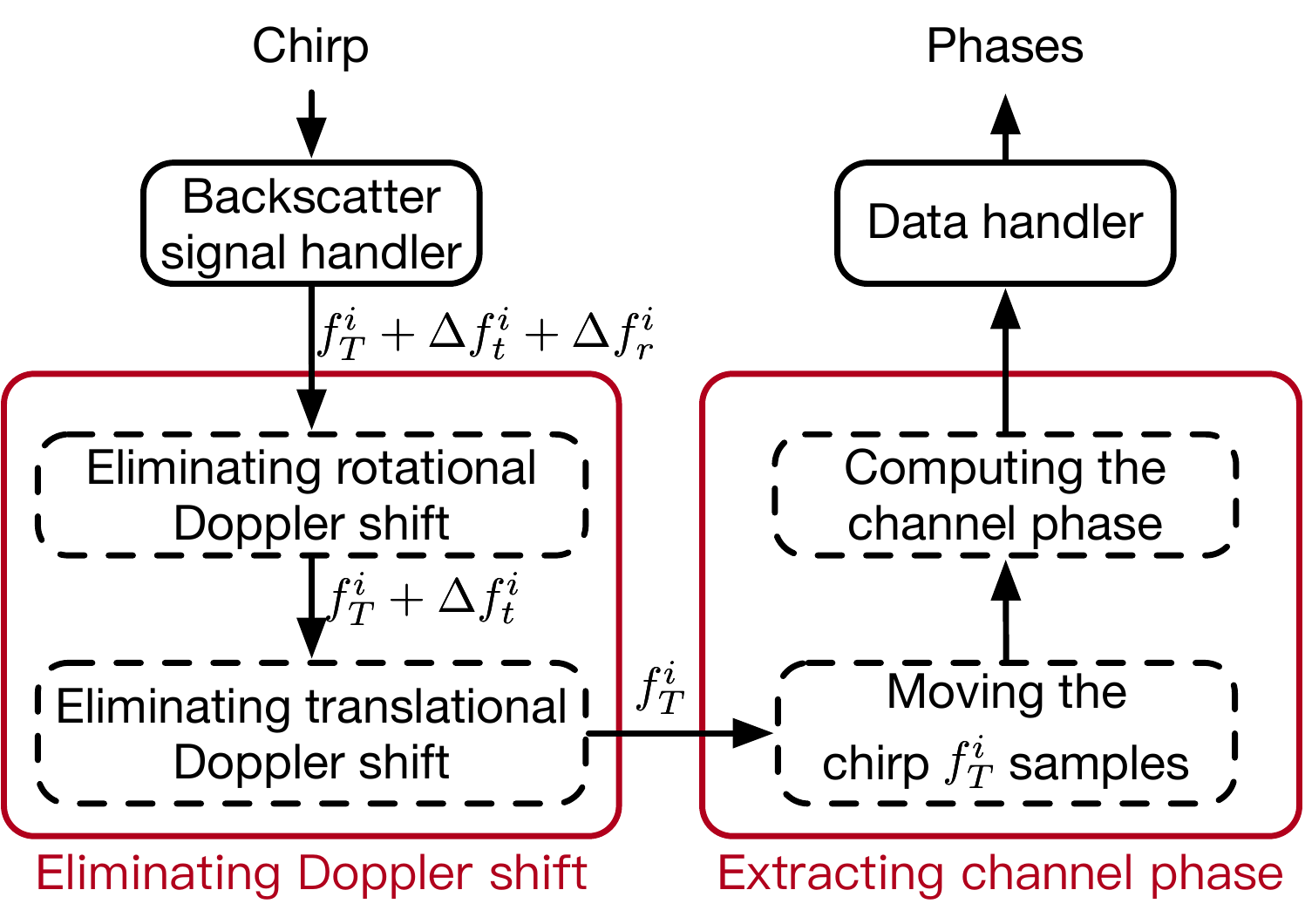}\vspace{-0.3cm}
		\caption{Phase extraction workflow.} \label{fig:phase_workflow}
	\end{minipage}
\end{figure}

\subsection{Below-Noise Channel Phase Extraction}
\label{subsec:phase}
Since MAVs are expect to carry out emergency tasks like fire rescue, the system desires the localizability with a single anchor (its controller) and without prior knowledge of the work space, being instantly deployable and operable wherever required. The position of a target referring to a single anchor can be represented by the angle $\phi$ and the range $r$ of the target to the anchor as polar coordinates. And both the parameters can be inferred by the channel phase of the signal. 

The channel phase extraction for chirp signals has two steps as shown in Fig.~\ref{fig:phase_workflow}: we first combat the Doppler effect to estimate the beginning of the chirp and then we extract the channel phase leveraging the linearity of the chirp frequencies. 

The channel phase extraction for chirp signals has two steps: we first combat the Doppler effect to estimate the beginning of the chirp and then we extract the channel phase leveraging the linearity of the chirp frequencies. 

To estimate the beginning of the chirp, we leverage a key property of the chirp signal: a time delay in the chirp signal translates to frequency shift. Ideally, decoding the original upchirp with a downchirp produces a peak in the first FFT bin (see Fig.~\ref{fig:overview}). When a tag is separated from the controller, the backscatter signal handler receives the signal with a timing offset of the signal's round trip. The peak appears in the shifted bin $f_s$. If we move the beginning of the received chirp $f_s$ samples closer to its real beginning and repeat the decoding operation, there will be a new peak at the first FFT bin again and the symbol at this instant is the beginning of the transmission. However, under the MAV's mobility, the signal additionally experiences the Doppler frequency shift. The shifted bin $f_s$ is a mixed result of the timing offset and the Doppler effect. The above operation can no longer recover the beginning of the chirp.

\begin{figure}
	\centering
	\begin{minipage}[b]{0.4\textwidth}\centering
		\center
		\includegraphics[width=1\textwidth]{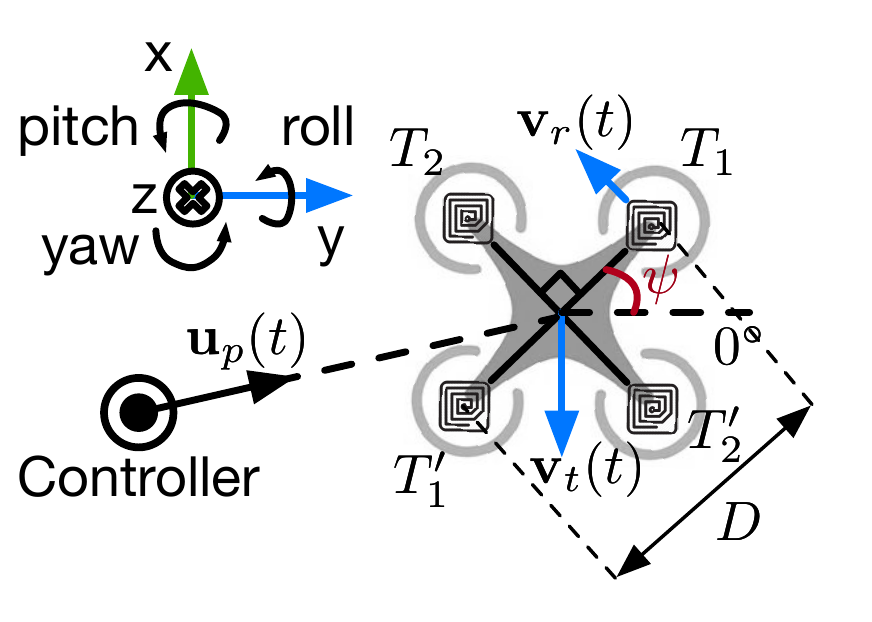}\vspace{-0.3cm}
		\caption{Motional states of the MAV.} \label{fig:tags}
	\end{minipage}
\end{figure}

Our solution leverages the kinetics and the structure of a MAV. We attach four backscatters on the landing gear of a MAV. As shown in Fig.~\ref{fig:tags}, the Doppler frequency shift of a tag, \eg, tag $T_1$, is a combinatorial result of translation and rotation. The shift $\Delta f(t)$ can be expressed as, 
\begin{equation}
	\Delta f(t) = \frac{f_c}{c} \mathbf{u}_p(t)\cdot\left[\mathbf{v}_t(t) + \mathbf{v}_r(t)\right] = \Delta f_t(t) + \Delta f_r(t),
	\label{eqn:doppler}
\end{equation}
where $\mathbf{u}_p(t)$ is the unit vector that represents the direction from the MAV to the controller, $f_c$ the carrier frequency, $c$ the speed of RF signals in the medium. $\mathbf{v}_t(t)$ and $\mathbf{v}_r(t)$ are the translational velocity and the rotational velocity. $\Delta f_t(t)$ and $\Delta f_r(t)$ corresponds to the translational shift and the rotational shift. To estimate the beginning of the chirp, we need to isolate the frequency shift translated from the timing offset by eliminating the effect of Doppler shift. 

\begin{figure}
	\centering
	\begin{minipage}[b]{0.4\textwidth}\centering
		\center
		\includegraphics[width=1\textwidth]{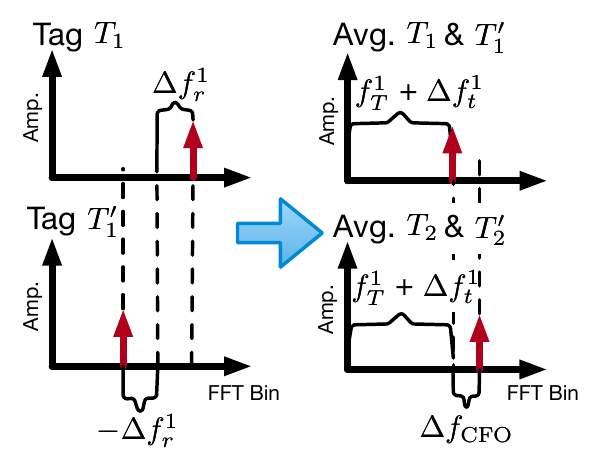}\vspace{-0.3cm}
		\caption{Rotational shift elimination.} \label{fig:peaks}
	\end{minipage}
\end{figure}

{\bf Eliminating the effect of Doppler shift}. 
We first eliminate the effect of the rotational shift by the key observation that any pair of opposing tags on the landing gear, \eg, tags $T_1$ and $T_1^\prime$ in Fig.~\ref{fig:tags}, always have rotational velocities with {\em the same magnitude but opposite directions} and all tags share {\em the same translational velocity}. Thus, averaging the shifted peak of two opposing tags eliminates the rotational shift as shown in Fig.~\ref{fig:peaks}. Specifically, decoding the backscattered signals from a pair of opposing tags, we obtain the FFT bin indices, $\hat{B}_i$ and $\hat{B}_i^\prime$,
\begin{equation}
	\hat{B}_i = f_T^i + \Delta f_t^i + \Delta f_r^i, \; \hat{B}_i^\prime = f_T^{i^\prime} + \Delta f_t^i - \Delta f_r^i,
	\label{eqn:shift}
\end{equation}
where $f_T^i$ and $f_T^{i^\prime}$ are the frequency shift translated by the timing offset, $\Delta f_t^i$ and $\Delta f_r^i$ the translation shift and the rotational shift of tag $i$. Note that $f_T^i \approx f_T^{i^\prime}$ since their maximal difference is the translated shift from the traveling time of the distance between a pair of opposing tags, \ie, the diameter $D$ of the MAV, which is negligible as $D$ ($66$ cm for the DJI M100) is too small for the speed of RF signal propagation. Thus, averaging them, \ie, $1/2(\hat{B}_i + \hat{B}_i^\prime) = f_T^i + \Delta f_t^i$, eliminates the rotational shift.

Since the two pairs of tags on the MAV are structurally symmetric, when we perform the above operation to each pair, the results are expected to be identical. However, they exhibit a slight difference as shown in Fig.~\ref{fig:peaks}. This is because the micro-controllers of the tags are not synchronized with the controller, it introduces an additional carrier frequency offset (CFO) for each tag, which is a constant. In our approach, $\Delta f_{\text{CFO}}$ is the difference of CFOs upon averaging the two pairs of tags, which is still a constant. We can simply apply this to the rest of the transmission to estimate the right chirp phase.

Now we eliminate the translational shift $\Delta f_t^i$ to isolate the frequency shift $f_T^i$ translated from the timing offset. Then, we can obtain the signal at the real beginning of the transmission by moving the beginning of the received chirp $f_T^i$ samples. $\Delta f_t^i$ can be tracked using the accelerations measured by the onboard IMU. Initially, the MAV is about to take off. At this initial stage, there is no motion, $1/2(\hat{B}_i + \hat{B}_i^\prime)$ is already the frequency shift $f_T^i$. Thus, the channel phase can be obtained according to the workflow (Fig.~\ref{fig:phase_workflow}). Then, we specify $\mathbf{u}_p(t)$ in Eqn.~\eqref{eqn:doppler} by our angle estimation algorithm in \cref{subsec:pose}. When the MAV takes off, the accelerations measured by IMU can track the translational velocity $\mathbf{v}_t(t)$. Thus, $\Delta f_t^i = f_c/c\cdot\mathbf{u}_p(t)\cdot\mathbf{v}_t(t)$ and $f_T^i = 1/2(\hat{B}_i + \hat{B}_i^\prime) - \Delta f_t^i$. Note that integrating the accelerations to obtain the velocity will suffer from the temporal drift. The super-accuracy algorithm in \cref{sec:pose} corrects the drift and feeds back to the controller. 


{\bf Extracting channel phase}.
At this stage, we have corrected the signal to the symbol at the beginning of the transmission. Now we compute the channel phases of all frequencies in the chirp. Taking an FFT to the multiplication of the corrected upchirp and the downchirp adds the phase across all the frequencies in the chirp. Assume that a linear chirp has $N$ frequencies and the channel path of the signal remains constant throughout the duration of the chirp, then the phase of the received chirp changes linearly with the frequency. We have
\begin{equation}
\hat{\theta}_\Sigma = \theta_1 + \theta_2 + \cdots + \theta_N = \theta_1 + \theta_1 \frac{f_2}{f_1} + \cdots + \theta_1 \frac{f_N}{f_1},
	\label{eqn:phase_sum}
\end{equation}
where $f_1, \cdots, f_N$ are explicitly defined when generating the chirp signal. Solving the above equation obtains the channel phases of all frequencies in the chirp.

Note that this method requires a short chirp duration to be within the channel coherent time. Meanwhile, the signal needs good decoding capability, which is proportional to the product of signal duration and bandwidth. To this end, we choose the parameters of CSS signals that conform to LoRa standard as $SF = 12$, $BW = 500$ KHz, and thus the chirp duration is $8$ ms.

\subsection{Below-Noise Pose Sensing}
\label{subsec:pose}

So far we have obtained the channel phases of backscatter signals. We now use them to estimate the pose feature, including the range and angle of a MAV to its controller for positioning, and the MAV's rotation for determining the orientation.

{\bf Range estimation}. 
Assume that the controller is separated from a tag on the MAV by a distance of $r$. A linear chirp signal with $N$ frequencies transmitted by the controller propagates a total distance of $2r$ for the round trip to and from the tag. The wireless channel of such a signal is, 
\begin{equation}
    \mathbf{H} = \left[\gamma_1 e^{-j2\pi f_1\frac{2r}{c}}, \gamma_2 e^{-j2\pi f_2\frac{2r}{c}}, \cdots, \gamma_N e^{-j2\pi f_N\frac{2r}{c}}\right],
    \label{eqn:range}
\end{equation}
where $\gamma_i$ is the attenuation corresponding to frequency $f_i$ in the chirp, $i = \{1, \cdots, N\}$. In the absence of multipath, we can use the obtained channel phases of the backscatter signal to estimate the range $r$. However, due to multipath, the obtained phases is actually the sum of phases of the direct-path signal and the multipath-reflected signals.

To combat multipath while conforming to LoRa protocol, we send multiple chirps in the channels of $900$ MHz band and combine the phase information across all these channels to simulate a wideband transmission. At a high level, a wideband signal can be used to disambiguate the multipath. There are $13$ channels separated by $2.16$ MHz with respect to the adjacent channels. We have four tags on the MAV which are configured to different frequency shifts for preventing the interference from the excitation signal. So, the controller can transmit excitation signals in $2$ channels and receive backscatter signals across $8$ channels. By combining them, the controller sends the phases at all the channels to the MAV through LoRa. Then, the MAV computes the range estimate by using an inverse FFT on the phases to get the time-domain multipath profile. We use a fixed energy threshold over this profile to identify the closest (most direct) path from the MAV. 

{\bf Angle estimation}. 
The angle of incident signals $\phi$ is also encoded in the phases of the signals. The backscattered chirp signal received by a linear array with $M$ antennas from $K$ propagation paths has the measurement matrix $\mathbf{X}$, 
\begin{equation}
	\begin{aligned}
		& \mathbf{X}  = \left[\mathbf{x}_1 \dotsc \mathbf{x}_N \right] = \mathbf{S}\left[\mathbf{F}_1 \dotsc \mathbf{F}_N \right], \\
		& \mathbf{S}\mathbf{F}_i = \left[ \mathbf{s}(\phi_1) \dotsc \mathbf{s}(\phi_K) \right]\left[ \gamma_{i1} \dotsc \gamma_{iK} \right]^\top, i = \{1, \cdots, N\},	\\
		& \mathbf{s}(\phi_k) = \left[ 1 \; e^{-j\eta \sin(\phi_k)} \dotsc e^{-j(M-1)\eta \sin(\phi_k)}  \right]^\top,
	\end{aligned}
	\label{eqn:angle}
\end{equation}
where $k = \{1, \cdots, K\}$, $\mathbf{F}_i$ denotes the attenuation factors of $K$ paths at frequency $i$ in the chirp, $\gamma_{ij}$ the attenuation factor of path $j$ at frequency $i$. $\mathbf{S}$ is the steering matrix where $s(\phi_k)$ denotes the steering vector of path $k$, and the constant $\eta = 2\pi d\frac{f_c}{c}$ where $d$ is the antenna spacing. $\phi_k$ is the angle of interest. We can see that the angle only exists in the steering matrix, contributing the phases in the complex elements of matrix $\mathbf{X}$. 

Thus, even without the attenuation information, we can use the obtained phases to construct a {\em virtual measurement matrix} of which all complex elements have unit attenuation with the phases of frequencies in the chirp to allow the angle estimation. The virtual measurement matrix $\hat{\mathbf{X}}$ can be written as
\begin{equation}
	\hat{\mathbf{X}} = 
	\begin{bmatrix}
		e^{j\theta_{11}} 	& e^{j\theta_{12}} 	& \cdots 	& e^{j\theta_{1N}}	\\
		e^{j\theta_{21}} 	& e^{j\theta_{22}} 	& \cdots 	& e^{j\theta_{2N}}	\\
		\vdots				& \vdots 				& \ddots	& \vdots				\\
		e^{j\theta_{M1}} & e^{j\theta_{M2}}	& \cdots 	& e^{j\theta_{MN}}
	\end{bmatrix},
\end{equation}
where $\theta_{ij}$ denotes the phase of antenna $i$ at frequency $j$. Applying $\hat{\mathbf{X}}$ to the super-resolution angle estimation technique~\cite{kotaru2015spotfi}, we obtain the direct-path angle of a tag to the controller. The four tags provide four angles for every chirp. We compute the harmonic mean of the four angles as the final result. 

By far, we model the angle estimation in 2D case (Eqn.~\eqref{eqn:angle}) for ease of presentation. It is trivial to be extended to 3D case. Specifically, the 3D angle is represented by two parameters, azimuth angle $\phi$ and elevation angle $\xi$. It can be expressed by a vector $\mathbf{a} = [\cos\phi\sin\xi,\;\sin\phi\sin\xi,\;\cos\xi]^\top$. Since the linear array is only capable of determining one parameter, we change it to a 2D circular array with the three antennas. This will slightly change the steering vector in Eqn.~\eqref{eqn:angle}. Then, we can jointly search the two parameters in a very efficient way~\cite{xie2019md} to estimate the 3D angle. Or we can save the two-dimensional search using a barometer, which is a common sensor equipped on MAVs, to measure the MAV height $h$. With the range $d$ between the MAV and its controller, the elevation angle is fixed by $\xi = \arccos\left(\frac{h}{d}\right)$. Then, the angle estimation in 3D reduces to the problem in 2D case described above.

{\bf Rotation estimation}. 
The real problem to determine a MAV's orientation is how to anchor the yaw, a.k.a., heading. The orientation can be represented by Euler angles: roll $\alpha$, pitch $\beta$, and yaw $\psi$ for a rotation around $x$, $y$, and $z$ axes (Fig.~\ref{fig:tags}). And it can be computed by integrating the 3D angular velocity readings from the onboard IMU. The results however suffer from temporal drifts due to the inherent noise of IMU. Thanks to IMU that it drifts in four degrees of freedom, which includes 3D position and yaw angle. When equipping an IMU on an aerial vehicle, the yaw typically indicates the heading of the vehicle. Therefore, we need drift-free rotation estimates to fix the heading.


Our idea is that the rotational shift is solely determined by the rotation. We can use it to map the rotation. According to Eqn.~\eqref{eqn:shift}, subtracting the indices of the peaks from two opposing tags $\hat{B}_i$ and $\hat{B}_i^\prime$ gives the rotational frequency shift,
\begin{equation}
	\Delta \hat{B}_i = \hat{B}_i - \hat{B}_i^\prime  = f_T^i - f_T^{i^\prime} + 2\times \Delta f_r^i \approx 2\times \Delta f_r^i.
	\label{eqn:rotational_shift}
\end{equation}

Now we model the rotational shift. We denote the angle of the MAV to its controller as $\phi$ and the MAV's rotation as $\psi$ (refer to Fig.~\ref{fig:tags}), then $\mathbf{u}_p = \left[\cos\phi \; \sin\phi \right]^\top, \; \mathbf{v}_r = \frac{D}{2}\omega\left[ \cos(\psi + \frac{\pi}{2}) \; \sin(\psi + \frac{\pi}{2}) \right]^\top$, where $\omega$ is the angular velocity during the rotation. The rotational shift can be expressed as
\begin{equation}
	\Delta f_r^i = \frac{f_c}{c} \mathbf{u}_p\cdot \mathbf{v}_r = \frac{f_cD}{2c} \omega \times \sin\left(\phi - \psi\right).
	\label{eqn:relative_shift}
\end{equation}
The controller computes $\Delta f_r^i$ by Eqn.~\eqref{eqn:rotational_shift} and sends it to the MAV. $\phi$ can be obtained by the angle estimation algorithm. The gyroscope in IMU measures angular velocity $\omega$. The rest parameters are known constants. Thus, rotation $\psi$ can be solved by Eqn.~\eqref{eqn:relative_shift}. From this model, we can see that when $\phi = \psi$, the frequency shift is $0$. This is when the direction of the rotational velocity is orthogonal to the direction of the MAV to the controller. In this case, it poses an ambiguity of the rotation that $\psi$ could be $\phi$ or $\phi + \pi$. It is trivial to eliminate this ambiguity by using the measurements from the other pair of tags on the landing gear, which is an orthogonal structure.

%% file: state_estimation.tex
So far, we have discussed how Marvel measures the pose feature, including the range, angle and rotation of a MAV to its controller, based on backscattered CSS signals. In this section, we enable accurate state estimation by taking advantage of the onboard IMU to jointly optimize the state with the backscatter-based pose sensing. 

Since the controller's location is unknown, solving the state estimation problem consists of estimating the MAV state over its trajectory and the controller's location. The controller is essentially a key feature to the map of the environment in which the MAV moves. This falls into the simultaneous localization and mapping (SLAM) problem domain. Solutions to the SLAM problem can be either filtering-based or graph-based approaches. Filtering-based approaches are considered to be more efficient in computation as they only estimate the current robot state and the map. Their main drawback is that fixing the linearization points early may lead to suboptimal results. On the contrary, graph-based approaches can achieve better performance via repetitively linearizing past robot states and multi-view constraints~\cite{lin2018autonomous, lu2018simultaneous}. We employ a graph-based optimization framework to solve our state estimation problem.


\subsection{Problem Formulation}
\label{subsec:formulation}
\begin{figure}
	\centering
	\begin{minipage}[b]{0.4\textwidth}\centering
		\center
		\includegraphics[width=1\textwidth]{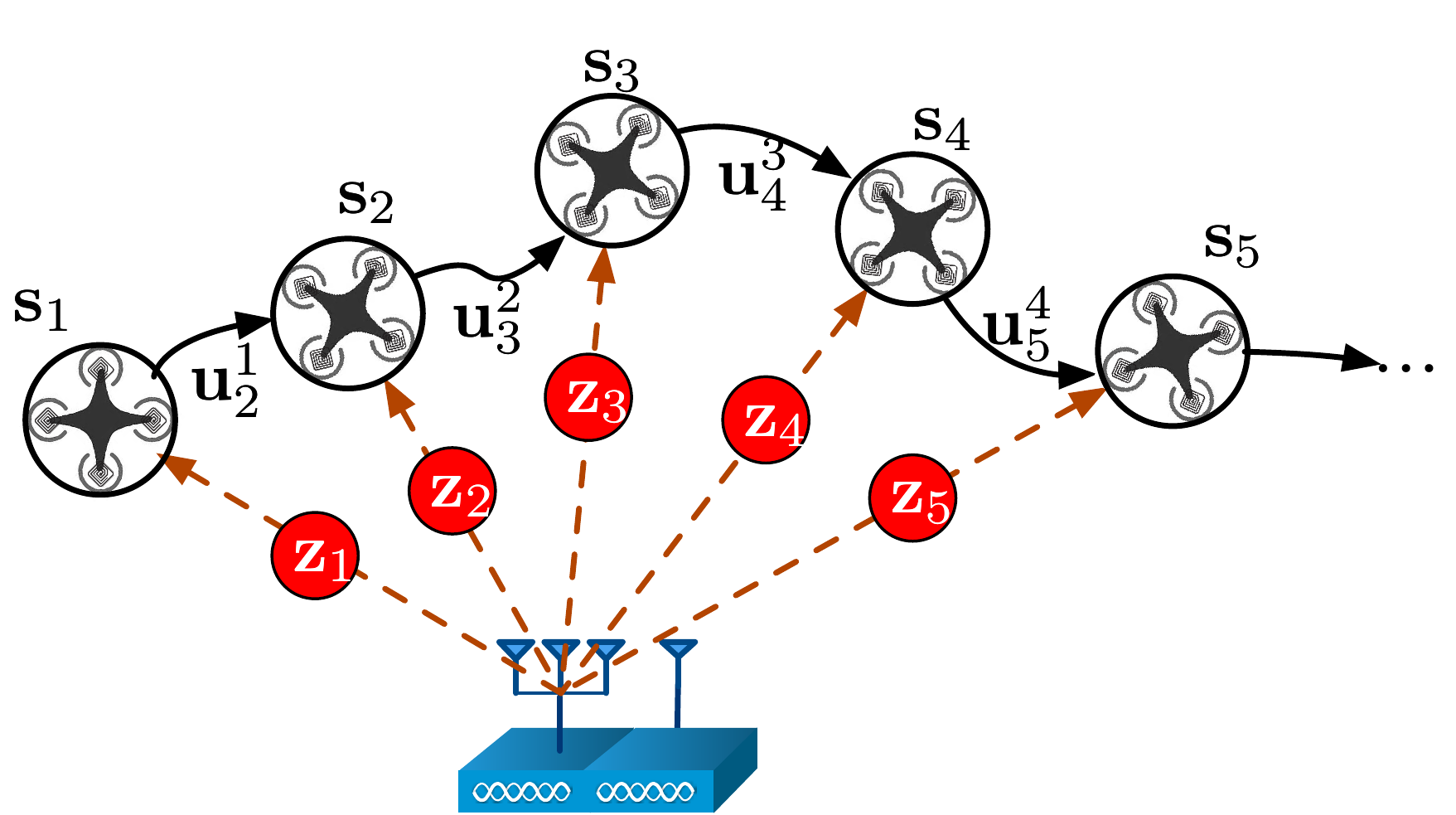}\vspace{-0.3cm}
		\caption{Graph-based optimization.} \label{fig:graphslam}
	\end{minipage}
\end{figure}

The graph representation of our state estimation problem is shown in Fig.~\ref{fig:graphslam}. Let $\mathbf{s}_k$ denote the state at time $k$. At each $k$, the MAV observes a set of backscatter sensing measurements $\mathbf{z}_k$ which include range $\hat{d}_k$, angle $\hat{\mathbf{a}}_k \in \mathbb{R}^{3}$ and yaw rotation $\hat{\psi}_k$. Typically, the IMU data rate ($100$ Hz) is much higher than the data rate of the backscatter sensing ($\approx 10$ Hz). Thus, there have been buffered multiple IMU measurements between two states. $\mathbf{u}_{k+1}^k$ is an integrated result over these measurements (specified in \cref{subsec:estimation}) that represents the odometry between two consecutive states, \eg, $\mathbf{s}_k$ and $\mathbf{s}_{k+1}$. 


To achieve real-time processing, we employ an incremental state update scheme~\cite{kaess2012isam2} that takes IMU and backscatter-based measurements in a fixed time interval for state estimation. As long as a new state with its backscatter-based measurements is available, our approach works in a sliding window fashion that incorporates the new state and removes the oldest state. The full state vector within the interval is defined as,
\begin{equation}
	\begin{aligned}
		\bm{\mathcal{S}} & = \left[ \mathbf{s}_0, \mathbf{s}_1, \cdots, \mathbf{s}_n, \bm{\rho} \right]	\\
		\mathbf{s}_k & = \left[ \mathbf{p}_k^0, \mathbf{v}_k^0, \mathbf{q}_k^0 \right], k \in [1, n], 
	\end{aligned}
\end{equation}
where $\mathbf{s}_k$ denotes $k$\spth state in the window, which contains position $\mathbf{p}_k^0$, velocity $\mathbf{v}_k^0$, and rotation $\mathbf{q}_k^0$ with respect to the first ($0$\spth) state. $\mathbf{q}_k^0 \in \mathbb{R}^4$ is the Hamilton quaternion~\cite{trawny2005indirect} representation of the rotation. We use the quaternion representation for modelling the odometry as a vector. $n$ is the number of states in the sliding window. $\bm{\rho}$ denotes the position of the controller.

Based on the state vector, we minimize the Mahalanobis norm of all measurement residuals to obtain a maximum a posteriori estimation: 
\begin{equation}
	 \min_{\bm{\mathcal{S}}} \left\{\sum_{j\in\mathcal{L}}\left\|\mathbf{e}_{\mathcal{L}}\left( \hat{\mathbf{z}}_j, \bm{\mathcal{S}} \right) \right\|_{\mathbf{P}_j}^2 + \sum_{k\in\mathcal{I}}\left\|\mathbf{e}_\mathcal{I}\left(\hat{\mathbf{u}}_{k+1}^k, \bm{\mathcal{S}} \right) \right\|_{\mathbf{P}_{k+1}^k}^2\right\},
	 \label{eqn:nonlinear}
\end{equation}
where $\mathbf{e}_{\mathcal{L}}\left( \hat{\mathbf{z}}_j, \bm{\mathcal{S}} \right)$ and $\mathbf{e}_\mathcal{I}\left(\hat{\mathbf{u}}_{k+1}^k, \bm{\mathcal{S}} \right)$ are measurement residuals for LoRa backscatter and IMU, respectively. $\mathcal{L}$ is the set of backscatter-based pose features and $\mathcal{I}$ denotes the set of IMU measurements. We choose the Mahalanobis norm to be the optimization objective because it takes into account the correlations of the data set. These correlations amongst internal states of different sensing modalities are key for any high-precision inertial-based autonomous system~\cite{leutenegger2015keyframe}. 

\subsection{Backscatter-inertial State Estimation}
\label{subsec:estimation}
We now solve the nonlinear system~\eqref{eqn:nonlinear} for state estimation via the Gauss-Newton algorithm. This involves linearizing the nonlinear system by the first order Taylor expansion of the residuals in~\eqref{eqn:nonlinear} around a reasonable initial guess. Since the backscatter-based pose sensing provides ranges, it is easy to obtain an initial guess of the state. 
The objective is to minimize the sum of the Mahalanobis norm of backscatter sensing and IMU residuals. The Mahalanobis norm in~\eqref{eqn:nonlinear} can be explicitly expressed as,
\begin{equation}
	\begin{aligned}
		\left\|\mathbf{e}_{\mathcal{L}}\left( \hat{\mathbf{z}}_j, \bm{\mathcal{S}} \right) \right\|_{\mathbf{P}_j}^2 & = {\mathbf{e}_{\mathcal{L}}^j}^\top \inv{(\mathbf{P}_j)} \mathbf{e}_{\mathcal{L}}^j	\\
		\left\|\mathbf{e}_\mathcal{I}\left(\hat{\mathbf{u}}_{k+1}^k, \bm{\mathcal{S}} \right) \right\|_{\mathbf{P}_{k+1}^k}^2 & = {\mathbf{e}_\mathcal{I}^k}^\top \inv{(\mathbf{P}_{k+1}^k)} \mathbf{e}_\mathcal{I}^k.
	\end{aligned}
\end{equation}
Here we briefly denote $\mathbf{e}_{\mathcal{L}}\left( \hat{\mathbf{z}}_j, \bm{\mathcal{S}} \right)$ and  $\mathbf{e}_\mathcal{I}\left(\hat{\mathbf{u}}_{k+1}^k, \bm{\mathcal{S}} \right)$ as $\mathbf{e}_{\mathcal{L}}^j$ and $\mathbf{e}_\mathcal{I}^k$. Next, we define the residuals and their corresponding covariance matrices.

{\bf Backscatter sensing residual}. 
Our backscatter-based pose sensing gives range $\hat{d}_j$, angle $\hat{\mathbf{a}}_j$, and yaw rotation $\hat{\psi}_j$. Rotation $\hat{\mathbf{q}}_j^0$ can be trivially derived by $\hat{\psi}_j$ with drift-free roll $\hat{\vartheta}_j$ and pitch $\hat{\varphi}_j$ provided by IMU. Given these measurements, the residual is defined as, 
\begin{equation}
	\mathbf{e}_{\mathcal{L}}^j = 
	\begin{bmatrix}
		\delta d_j					\\
		\delta\mathbf{a}_j	\\
		\delta\bm{\theta}_j
	\end{bmatrix} = 
	\begin{bmatrix}
		\left\|\hat{d}_j^2 -  \left( \mathbf{p}_j^0 - \bm{\rho} \right)^\top \left( \mathbf{p}_j^0 - \bm{\rho} \right)\right\|	\\
		\hat{\mathbf{a}}_j \times \left( \mathbf{p}_j^0 - \bm{\rho} \right)	\\
		2\left[\inv{(\hat{\mathbf{q}}_j^0)}\otimes \mathbf{q}_j^0 \right]_{ijk}
	\end{bmatrix},
\end{equation}
where $[\cdot]_{ijk}$ extracts the vector part of the quaternion, which is the approximation of the error-state representation. $\delta\bm{\theta}_j$ is the 3D error-state representation of quaternion. The covariance matrix $\mathbf{P}_j$ is the measurement noise matrix, which can be estimated by statistically analyzing the pose features.


{\bf IMU residual}. 
Based on the kinematics, the residual of IMU measurements can be defined as,
\begin{equation}
		\mathbf{e}_\mathcal{I}^k = 
		\begin{bmatrix}
			R(\mathbf{q}_0^k)\left( \mathbf{p}_{k+1}^0 - \mathbf{p}_k^0 + \frac{1}{2}\mathbf{g}^0\Delta t_k^2 \right) - \mathbf{v}_k^k\Delta t_k - \hat{\bm{\alpha}}_{k+1}^k	\\
			R(\mathbf{q}_0^k)\left( R(\mathbf{q}_{k+1}^0)\mathbf{v}_{k+1}^{k+1} + \mathbf{g}^0\Delta t_k \right) - \mathbf{v}_k^k - \hat{\bm{\beta}}_{k+1}^k	\\
			2\left[ \inv{(\mathbf{q}_k^0)} \otimes \mathbf{q}_{k+1}^0 \otimes \inv{(\hat{\bm{\gamma}}_{k+1}^k)} \right]_{ijk}
		\end{bmatrix},
\end{equation}
where $\hat{\mathbf{u}}_{k+1}^k = \left[\hat{\bm{\alpha}}_{k+1}^k, \hat{\bm{\beta}}_{k+1}^k, \hat{\bm{\gamma}}_{k+1}^k \right]^\top$ is the preintegrated result~\cite{lupton2012visual} using accelerations $\hat{\mathbf{c}}_t$ and angular velocities $\hat{\bm{\omega}}_t$, which are raw measurements provided by IMU at time $t$ within the time interval $\Delta t_k$ between two consecutive states. Specifically, $\hat{\bm{\alpha}}_{k+1}^k = \iint_{t\in[k, k+1]}R(\mathbf{q}_t^{k})\hat{\mathbf{c}}_t\,\mathrm{d}t^2$, $\hat{\bm{\beta}}_{k+1}^k = \int_{t\in[k, k+1]}R(\mathbf{q}_t^{k})\hat{\mathbf{c}}_t\,\mathrm{d}t$, $\hat{\bm{\gamma}}_{k+1}^k = \int_{t\in[k, k+1]}\bm{\gamma}_t^k \otimes \left[0\; \frac{1}{2}\hat{\bm{\omega}}_t\right]^\top \mathrm{d}t$ where $\otimes$ denotes the quaternion multiplication operation. $R(\mathbf{q}_t^k) \in \text{SO}(3)$ the conversion from the quaternion to the rotation matrix. The covariance $\mathbf{P}_{k+1}^k$ can be computed recursively by first-order discrete-time propagation within $\Delta t_k$, referring to~\cite{qin2017vins} for more details.



At this stage, all the terms of the nonlinear system~\eqref{eqn:nonlinear} have been explicitly defined. We then use Ceres Solver~\cite{ceres-solver} to solve the nonlinear problem.

%% file: eval.tex
\subsection{Implementation and Evaluation Methodology}

The controller is built by two colocated NI USRP-2943 nodes, each with a UBX160 daughterboard. They have four channels to be configured as a data handler with one antenna and a backscatter signal handler with three antennas. We configure USRP to work on $900$ MHz band and all signal parameters conform to LoRa standard. The three antennas for the backscatter signal handler are mounted to an acrylic pole separated by a distance of $16$ cm. The USRP nodes are synchronized using an external clock and frequency reference. We run the CSS decoding and the channel phase extraction on the controller. 

\begin{figure}
	\centering
	\begin{minipage}[b]{0.48\textwidth}\centering
		\center
		\includegraphics[width=1\textwidth]{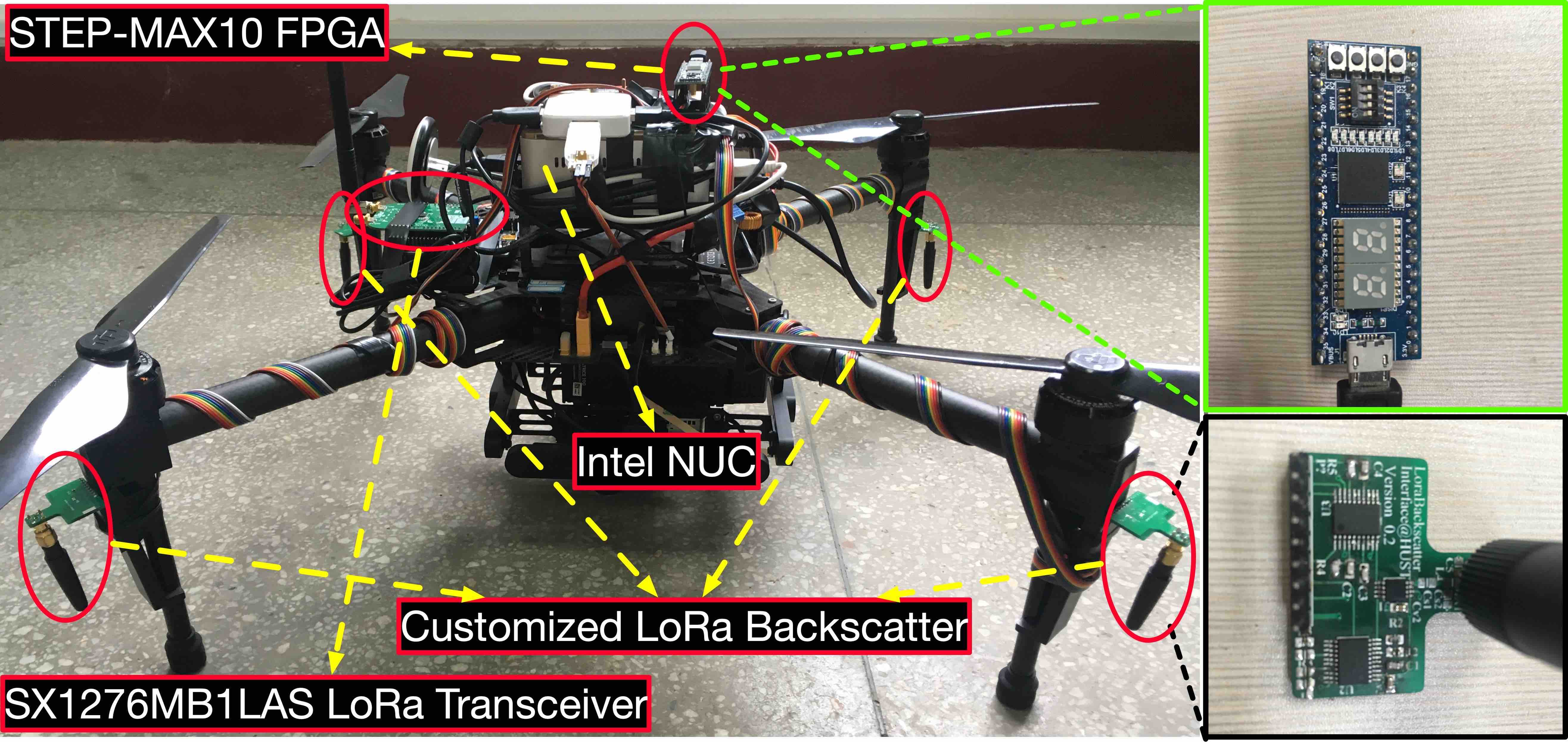}\vspace{-0.3cm}
		\caption{Experiment platform.} \label{fig:sys}
	\end{minipage}
\end{figure}

The MAV system is built by attaching an Intel NUC, a LORD MicroStrain 3DM-GX4-45 IMU, and an SX1276MB1LAS long-range transceiver on the DJI Matrice 100. In addition, there are four customized LoRa backscatter tags attached on the landing gear of the MAV. The backscatter uses the ADG919 and ADG904 RF switches to enable backscatter communications. The four backscatters are controlled by an Altera STEP-MAX10 FPGA. It configured them to shift $1$ MHz frequency with each other when backscattering the linear chirps with $500$ KHz bandwidth. We run Marvel on the Intel NUC with a 1.3 GHz Core i5 processor with $4$ cores, an $8$ GB RAM and a $120$ GB SSD, running Ubuntu Linux. The backscatter-based pose sensing module and the backscatter-inertial super-accuracy state estimation algorithm are written in C++. We use Robot Operating System (ROS) to be the interfacing robotics middleware. The experimental platform is shown in Figure~\ref{fig:sys}. 

We conduct experiments in both outdoors and indoors for the evaluations in long-range and through-wall settings. The outdoor experiments are conducted in an open field in front of an office building. There is no obstacle between the MAV and the controller. The indoor experiments are conducted in a MAV test site of $12\times 8$ square meters. The site is located on the basement level of an office building as shown in Fig.~\ref{fig:wall_setup}. Multiple rooms are separated by concrete walls, drywall, and wooden doors and have office furniture including tables, chairs, and computers. In the following, we first examine Marvel's sub-module accuracy and latency in micro-benchmark evaluation, and then we evaluate its system-level performance under different motions and environments.

\subsection{Micro-benchmark Evaluation}
\label{subsec:microbenchmark}
\begin{figure}
	\centering
	\begin{minipage}[b]{0.22\textwidth}\centering
		\center
		\includegraphics[width=1\textwidth]{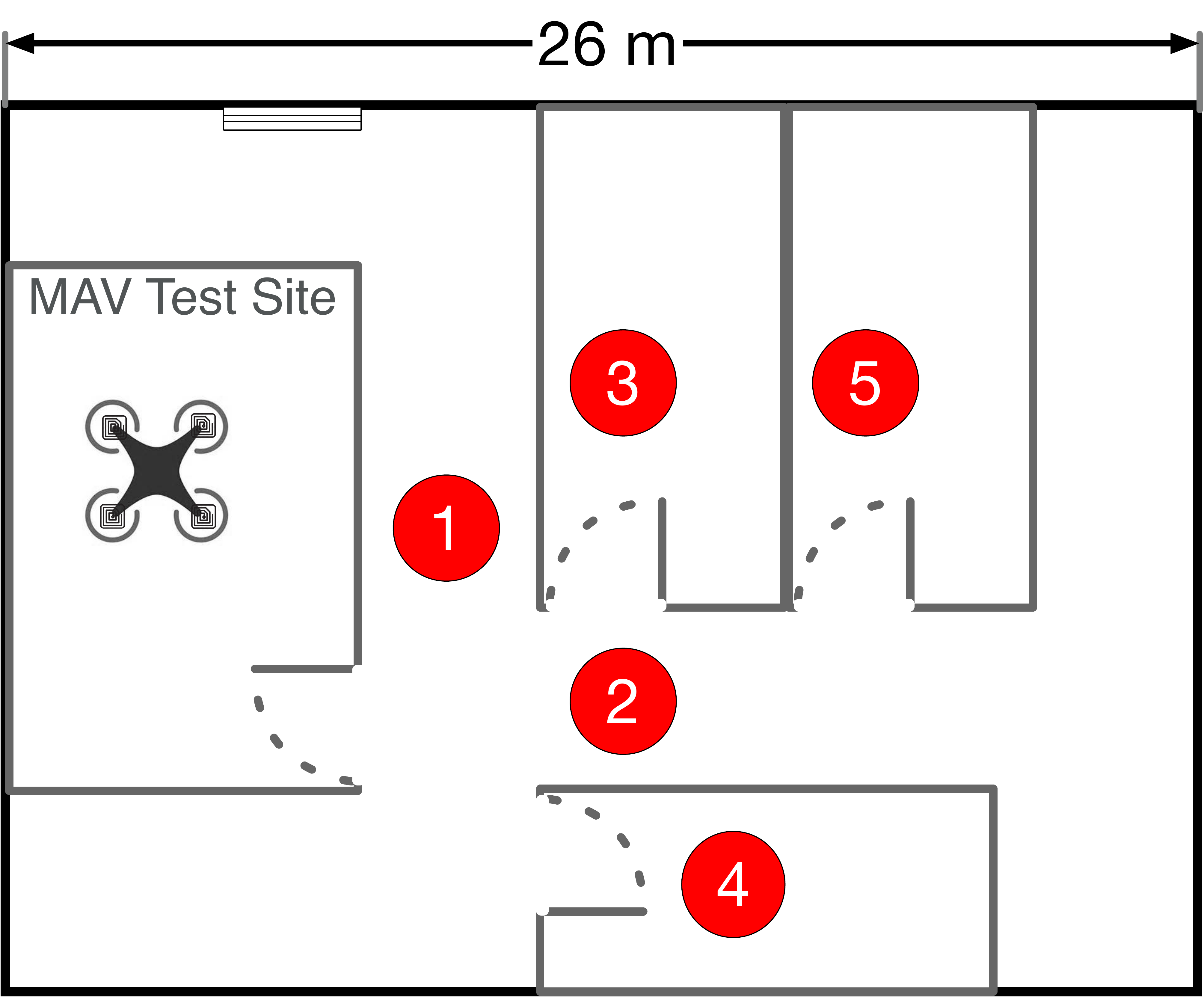}\vspace{-0.3cm}
		\caption{Through-wall setup.} \label{fig:wall_setup}
	\end{minipage}
	\hspace{-0.1cm}
	\begin{minipage}[b]{0.26\textwidth}\centering
		\center
		\includegraphics[width=1\textwidth]{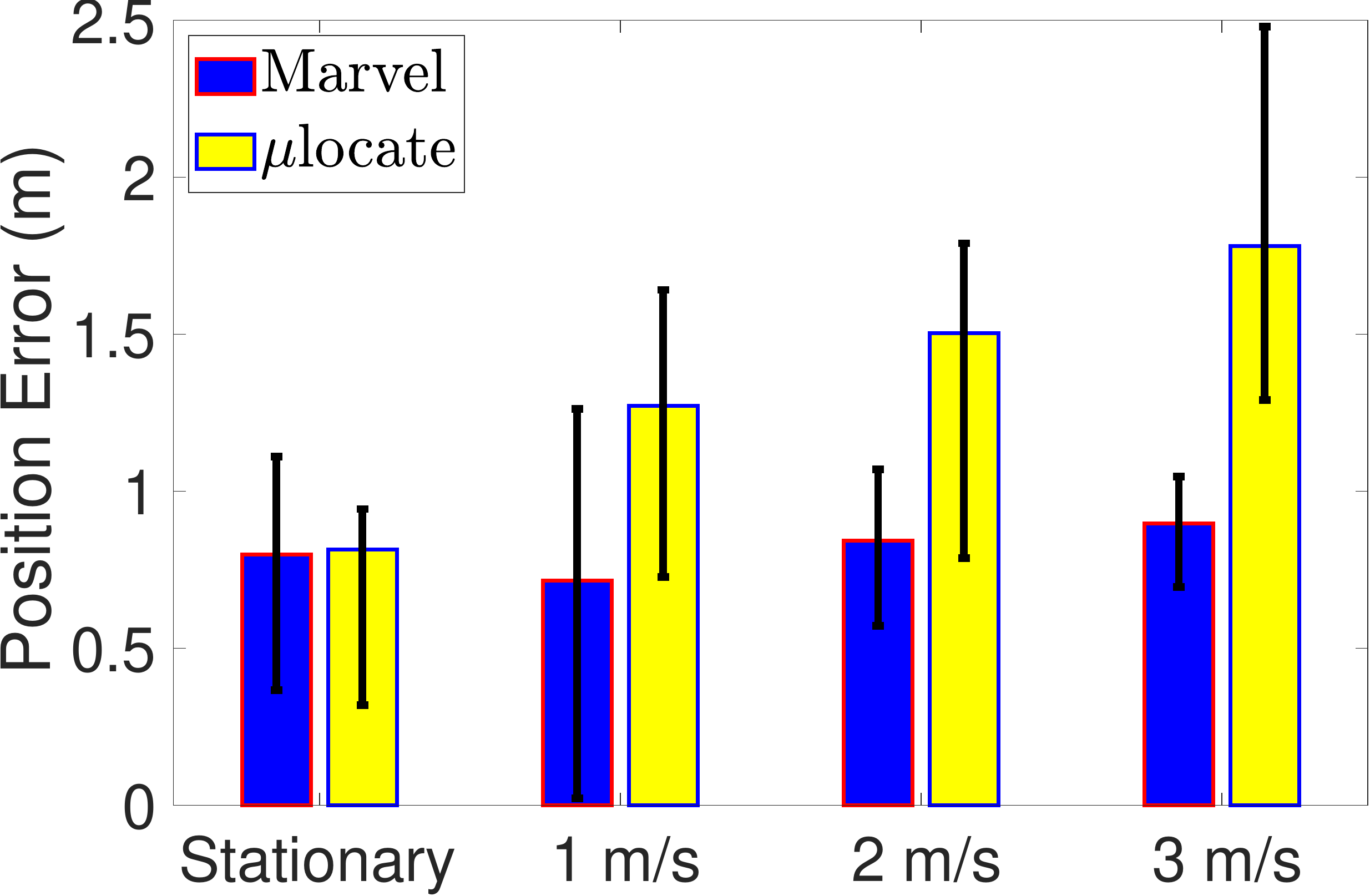}\vspace{-0.3cm}
		\caption{Positioning vs. speed.} \label{fig:comparison}
	\end{minipage}
\end{figure}

We evaluate the performance of positioning and rotation estimation, respectively. To evaluate the positioning approach, we build a sliding rail by the stepper motor ROB-09238~\cite{steppermotor} that supports the moving with a controllable speed. We place the MAV on a plate mounted on this rail. To evaluate the rotation estimation, we place the MAV on a plate mounted on the stepper motor and control the rotating speed. In long-range experiments, we place the controller at one end of the field and move the MAV away from the controller in increments of $10$ m. In through-wall experiments, we place the MAV in the test site and move the controller to different rooms (Fig.~\ref{fig:wall_setup}). There are three concrete walls between the controller and the MAV at location $5$. At each location, we repeat the experiment multiple times and compute the errors. 

\begin{figure}
	\centering
	\begin{minipage}[b]{0.23\textwidth}\centering
		\center
		\includegraphics[width=1\textwidth]{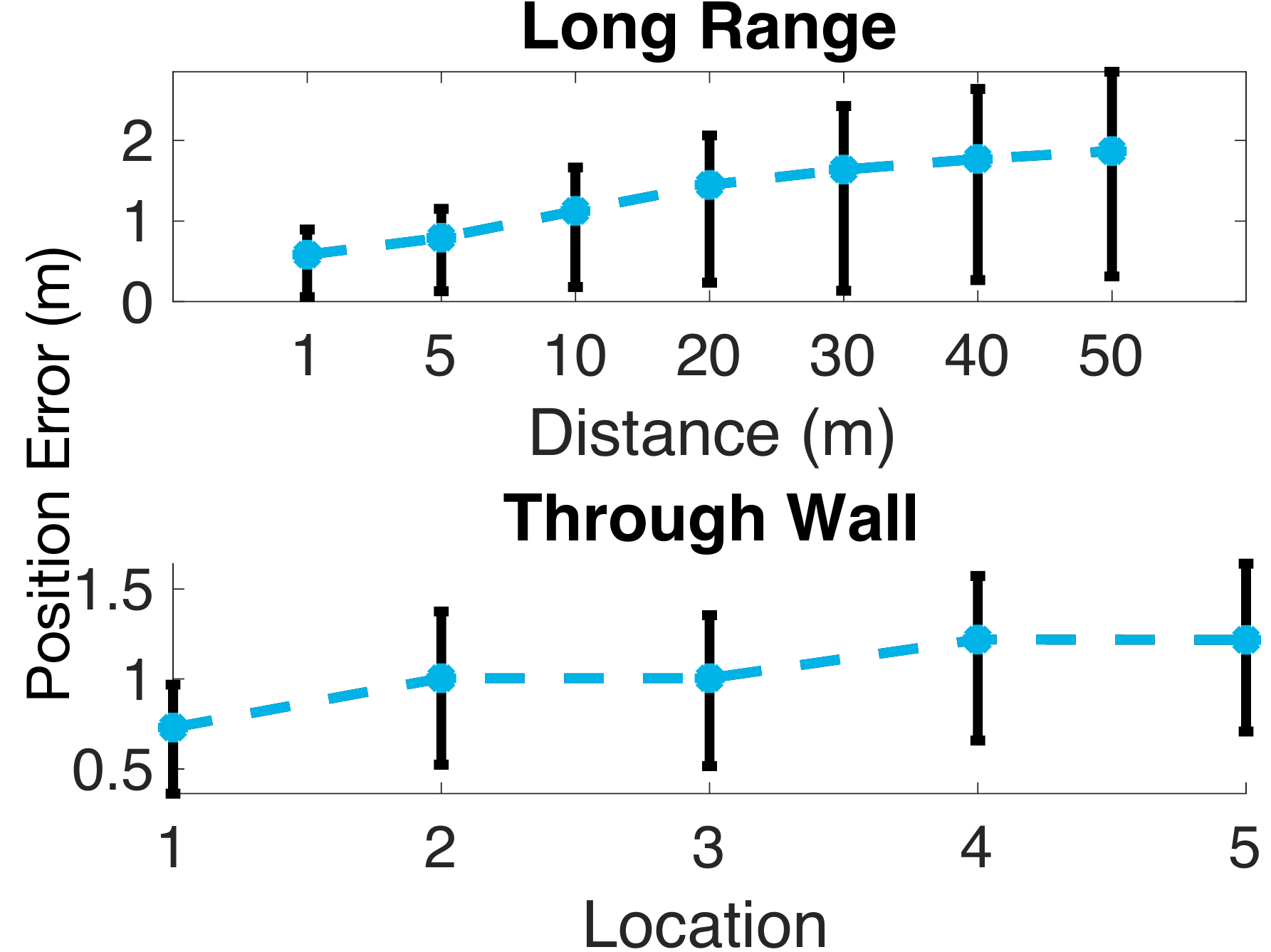}\vspace{-0.3cm}
		\caption{Positioning vs. setting.} \label{fig:position_error}
	\end{minipage}
	\hspace{-0.1cm}
	\begin{minipage}[b]{0.23\textwidth}\centering
		\center
		\includegraphics[width=1\textwidth]{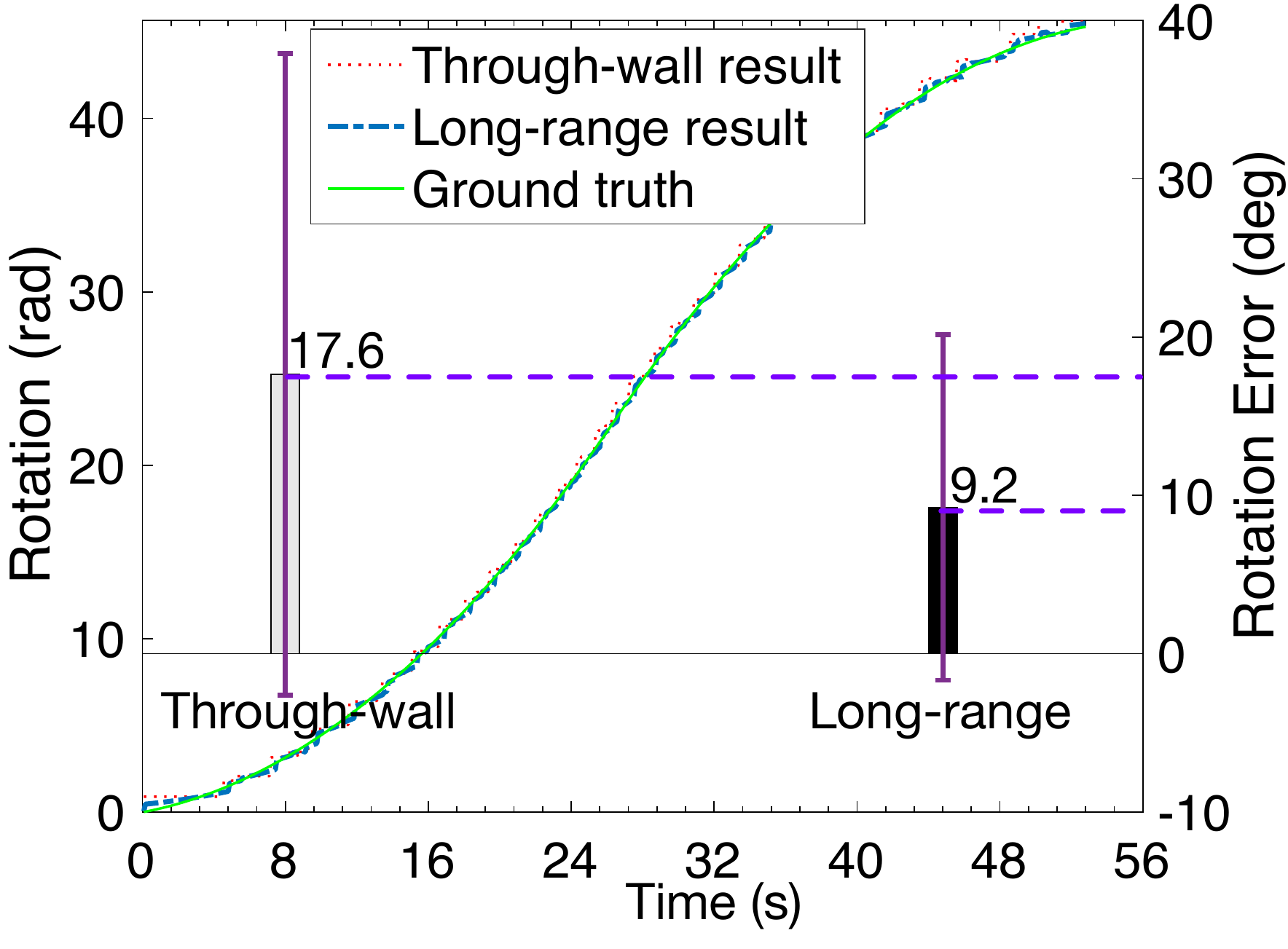}\vspace{-0.3cm}
		\caption{Rotation accuracy.} \label{fig:rotation_accuracy}
	\end{minipage}
\end{figure}

{\bf Positioning accuracy}. 
We first validate the positioning capability of Marvel in different speeds. We compare Marvel with the state-of-the-art CSS-based localization system, $\mu$locate~\cite{nandakumar20183d}, which operates correctly in semi-stationary scenarios. As shown in Fig.~\ref{fig:comparison}, the accuracies of the two approaches are similar in stationary case, whose mean error is around $0.8$ m. However, the error of $\mu$locate scales with the speed since its channel phase estimates are distorted by the Doppler frequency shift. Its position error reaches $2.45$ m in the worse case while Marvel's accuracy keeps steady. 

The positioning results in different settings are shown in Fig.~\ref{fig:position_error}. To demonstrate that our approach is resilient to the Doppler effect under mobility, we move the MAV along the rail in a speed of $3$ m/s, which is the maximum speed allowed. 

The long-range result shows that the error scales with the MAV-controller distance. Specifically, the position error of $0.58$ m at a distance of $1$ m, which increases to $0.79$ m at a distance of $5$ m. This further increases to $1.44$ m at a distance of $20$ m. This is due to the fact that the angle estimate with limited accuracy maps to a growing uncertain area of the MAV's position with the increasing distance. Our customized backscatter works at most $50$ m at which the worst case position accuracy is $2.66$ m. Beyond that distance, the power of the received signal is too low to decode even with the CSS coding. 

The through-wall result shows that the accuracies at different locations are similar because the MAV-controller distance does not vary much. But the accuracy in indoors is worse than at a distance of $1$ m in the open space due to the multipath fading. The worst case accuracy at location $5$ where has three walls blocking the MAV and the controller is $1.22$ m. Our controller is unable to receive the backscatter signal when it goes through more than three walls. 

In summary, the position accuracy is limited to meter level in both outdoors and indoors due to the limited signal bandwidth at the $900$ MHz band that we use. Nevertheless, with the aid of IMU, Marvel achieves decimeter-level accuracy as shown in \cref{subsec:systemlevel}.

{\bf Rotation estimation accuracy}. 
We evaluate the rotation estimation by controlling the stepper motor whose angular velocity starts from $0.2$ rad/s and increases by the rate $0.05$ rad/s until $1.5$ rad/s, and then decreases by the same rate to be back at $0.2$ rad/s. The whole process takes $52$ seconds as shown in Fig.~\ref{fig:rotation_accuracy}. We repeat the experiment $30$ times and analyze the data. As expected, the result in the through-wall setting is worse (mean error $17.6\degree$, standard deviation $20.3\degree$) than the other (mean error $9.2\degree$, standard deviation $10.9\degree$) due to the larger error of angle estimation in the presence of multipath. Fig.~\ref{fig:rotation_accuracy} also shows that our rotation estimation algorithm succeeds in closely tracking the MAV's rotation with varying angular velocities in both settings, providing drift-free results. 

\begin{figure}[t!]
    \centering
    \shortstack{
            \includegraphics[width=0.16\textwidth]{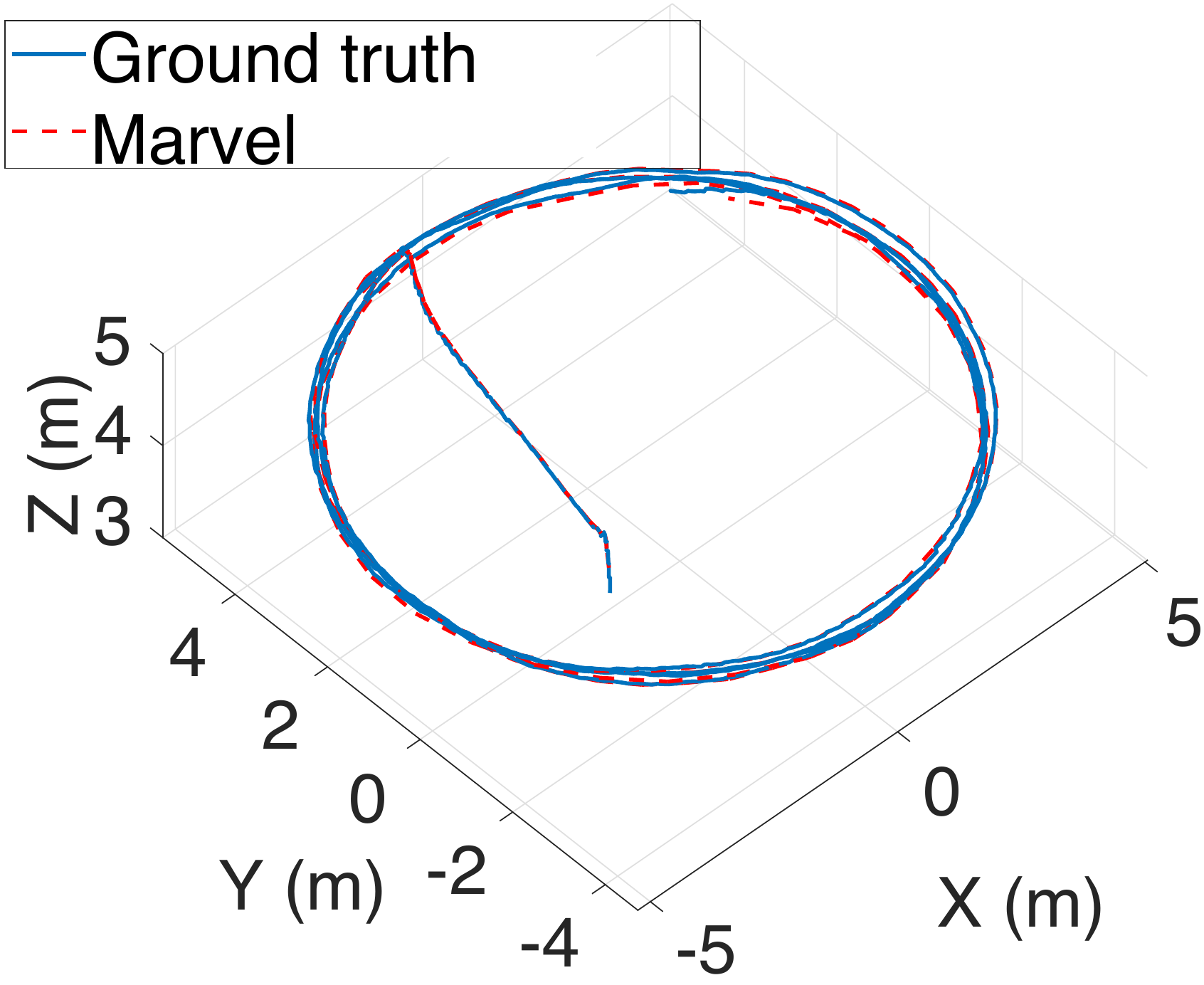}\\
            {\footnotesize (a) Circular trajectory}
    }\quad
    \shortstack{
            \includegraphics[width=0.15\textwidth]{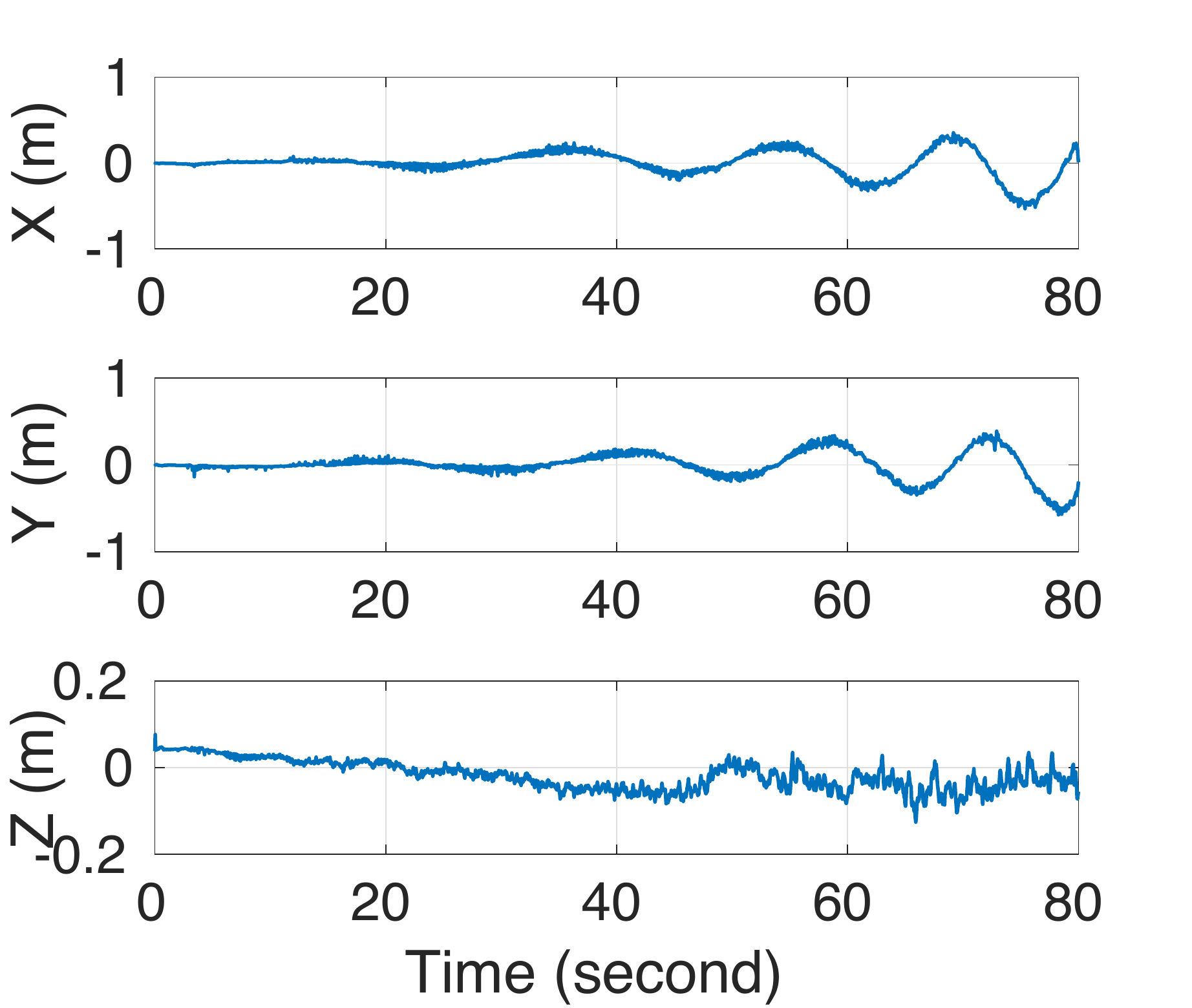}\\
            {\footnotesize (b) Position error}
    }
    \shortstack{
            \includegraphics[width=0.15\textwidth]{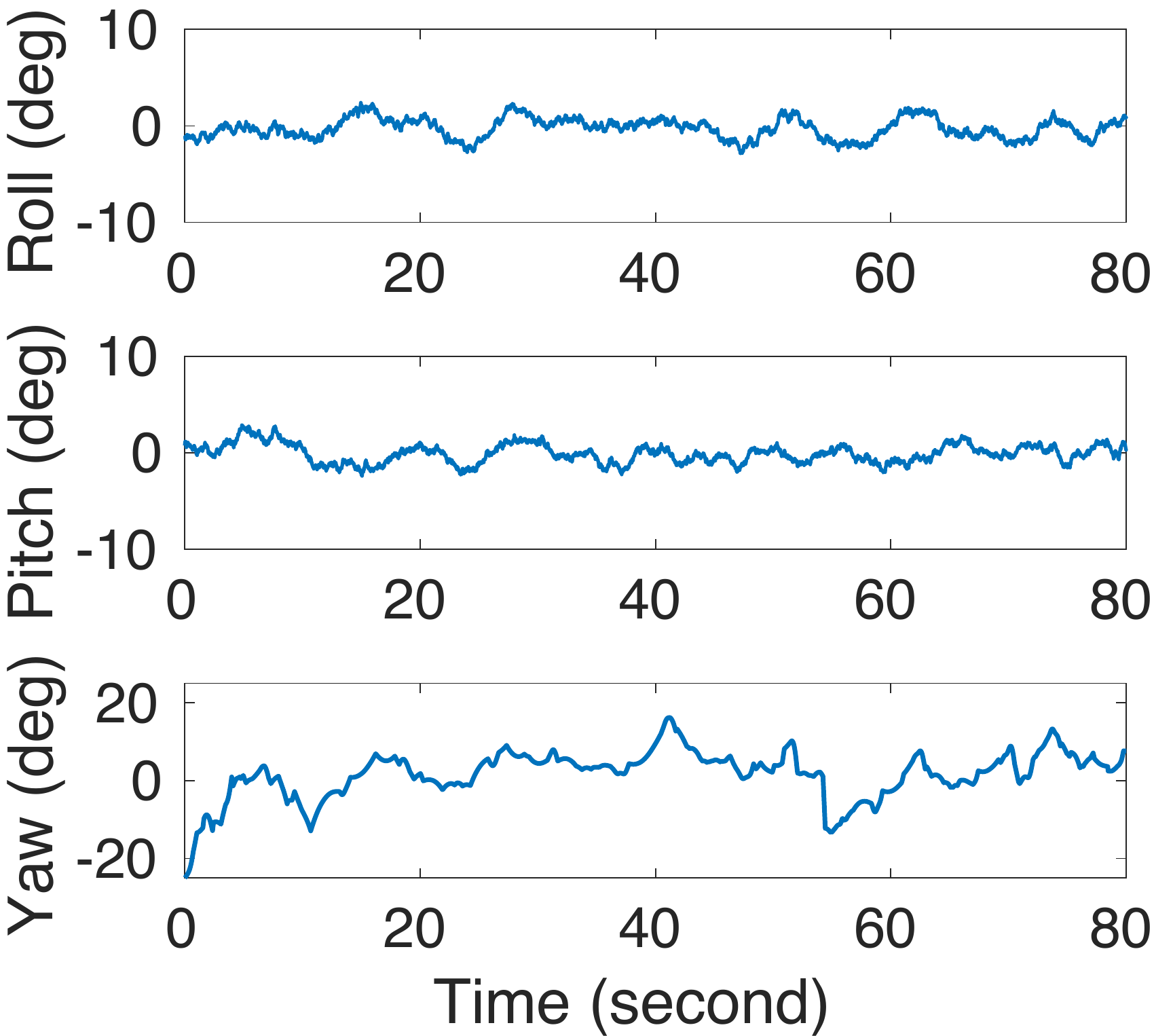}\\
            {\footnotesize (c) Orientation error}
    }
    \caption{Long-range state estimation.}
    \label{fig:losflight} 
\end{figure}

\subsection{System-level State Estimation}
\label{subsec:systemlevel}
We program the MAV to fly in different trajectories for evaluating the overall performance of Marvel in various motional patterns. The ground truth of the flight trajectories is provided by OptiTrack~\cite{optitrack}. The maximum linear velocity reaches $2.53$ m/s in this experiment. We skip the plot of linear velocity estimation as it is the first derivative of position and thus it is highly related to the position accuracy. 

 
In long-range experiments, the MAV flied in a circular trajectory. Since the backscattered signal cannot be decoded when the distance is longer than $50$ m, the controller is placed $20$ m away from the MAV before taking off to ensure that the MAV cannot go beyond the distance limitation during the flight. As shown in Fig.~\ref{fig:losflight}, the average error of state estimation is $33.66$ cm for positioning and $4.99\degree$ for orientation estimation. This demonstrates that the super-accuracy algorithm significantly increases the accuracy of pose tracking, enabling accurate state estimation. 

\begin{figure}[t!]
    \centering
    \shortstack{
            \includegraphics[width=0.16\textwidth]{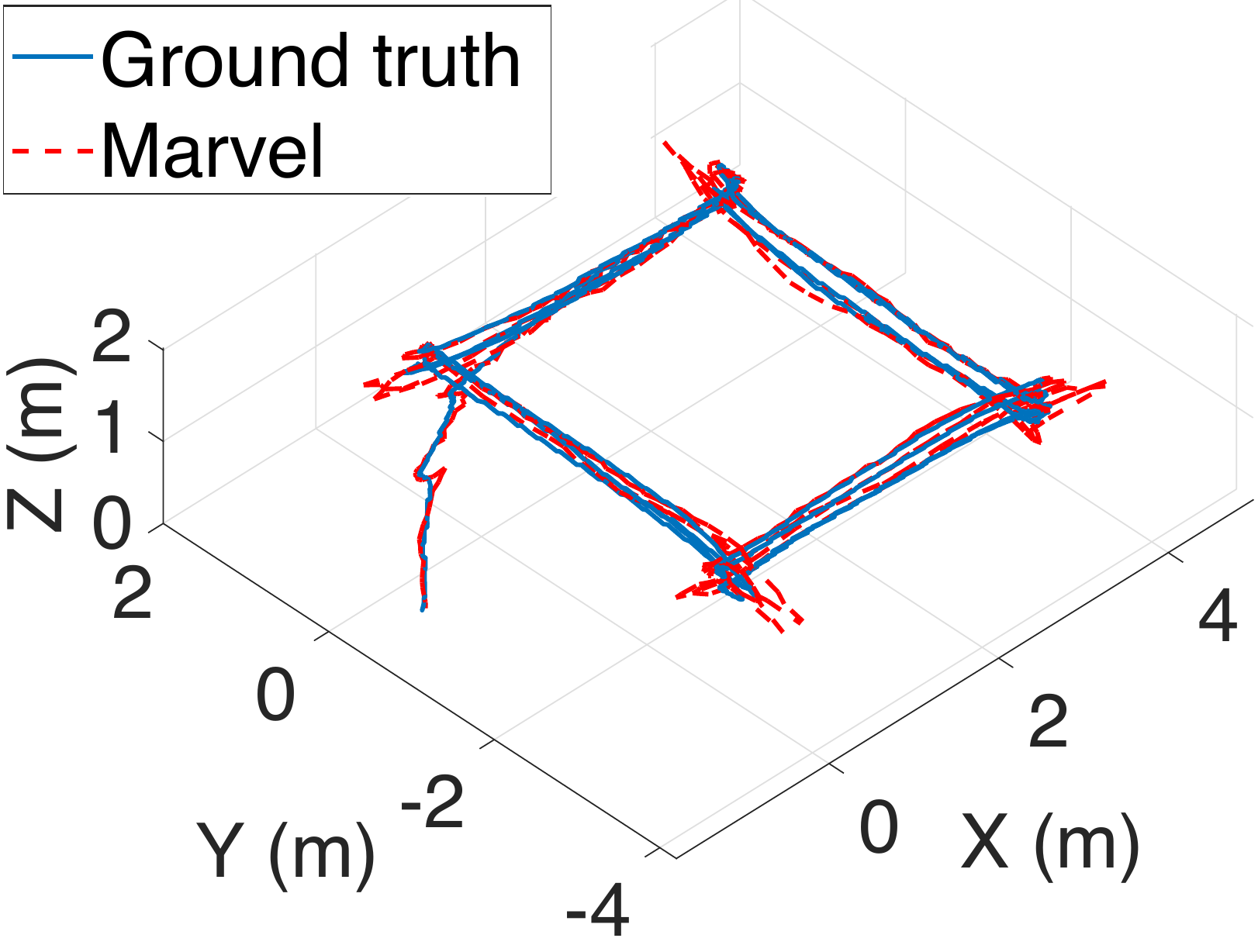}\\
            {\footnotesize (a) Square trajectory}
    }\quad
    \shortstack{
            \includegraphics[width=0.15\textwidth]{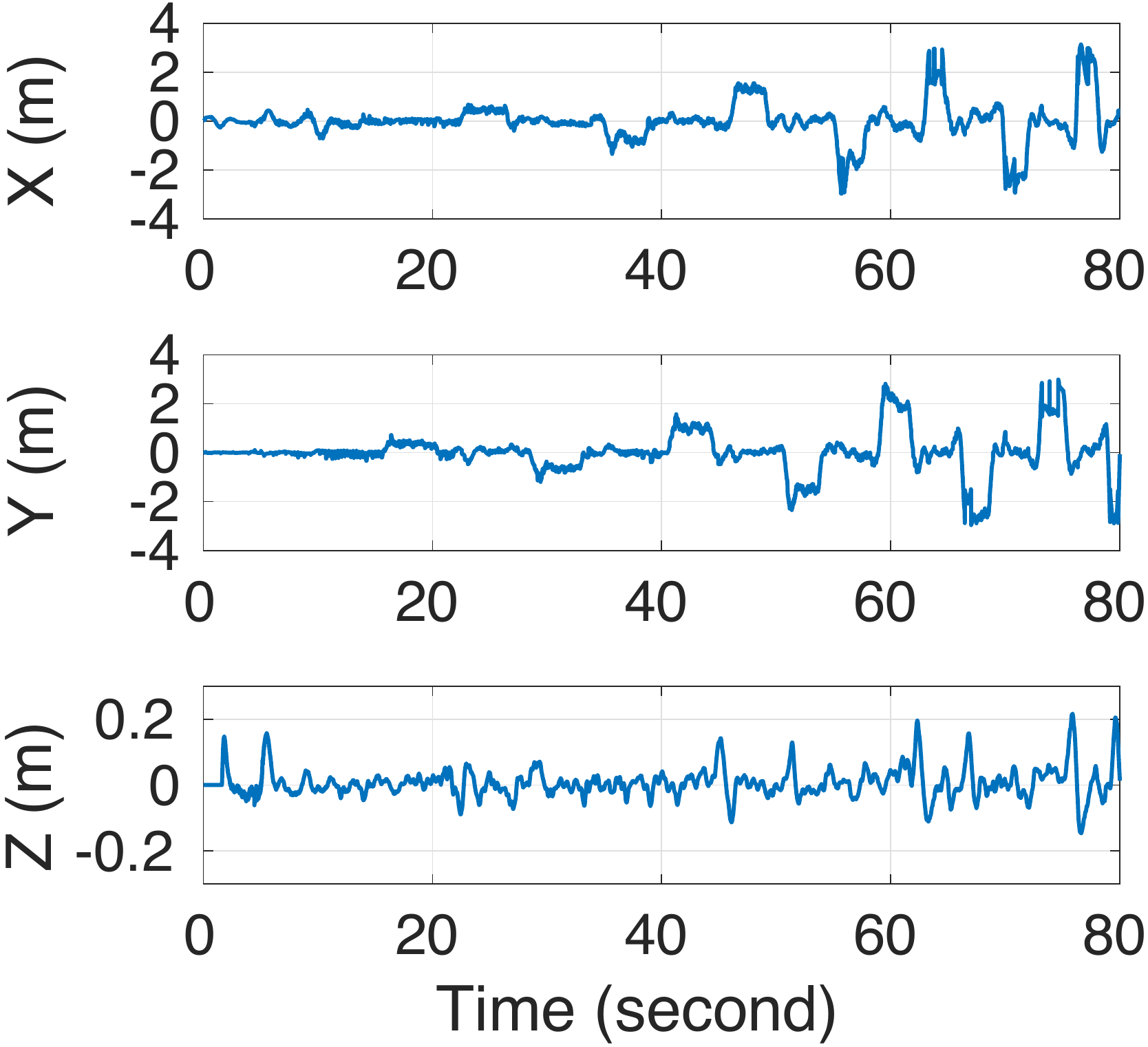}\\
            {\footnotesize (b) Position error}
    }
    \shortstack{
            \includegraphics[width=0.15\textwidth]{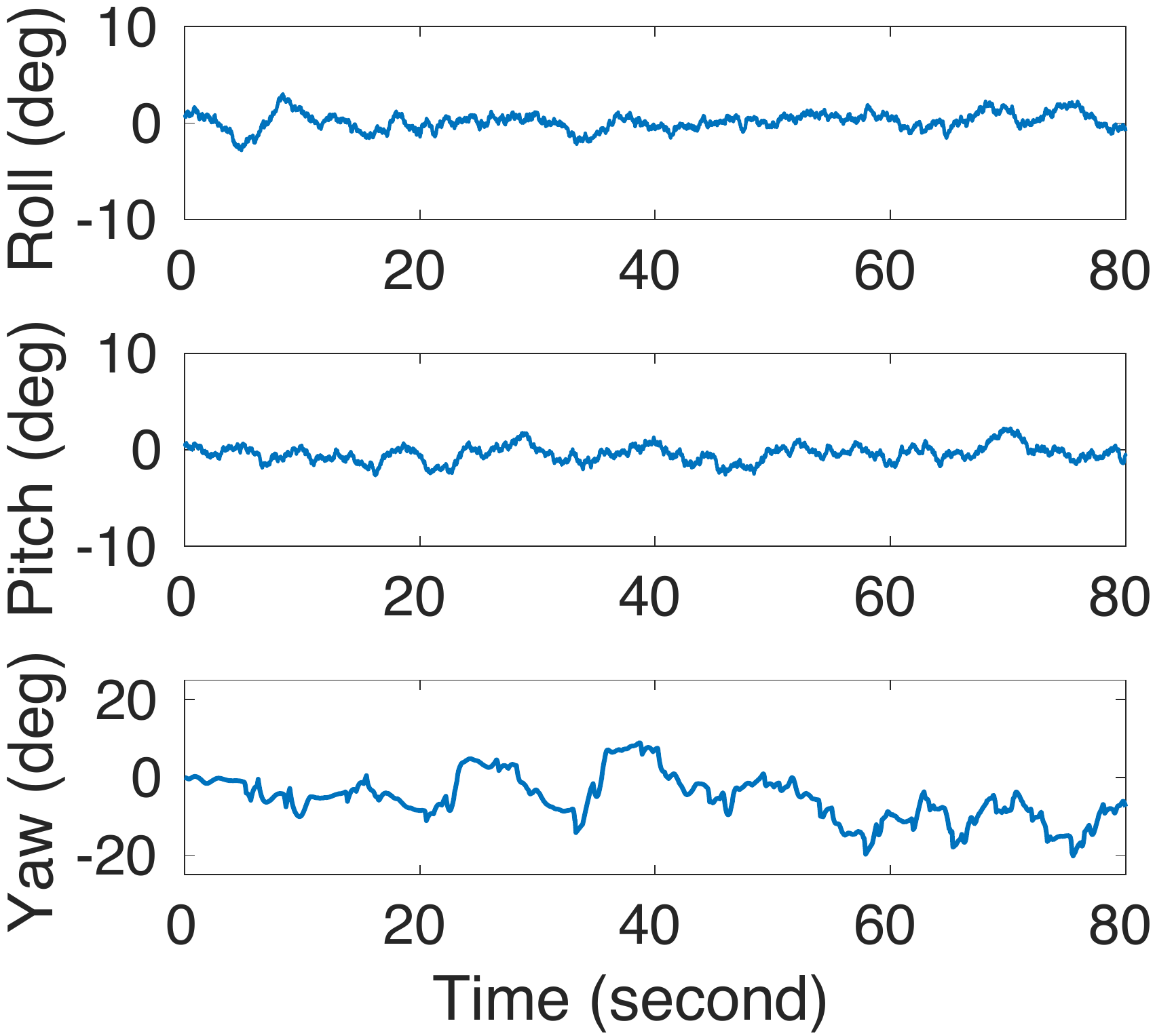}\\
            {\footnotesize (c) Orientation error}
    }
    \caption{Through-wall state estimation.}
    \label{fig:nlosflight}
\end{figure}

In through-wall experiments, for safety reasons, the MAV has to fly in the test site. We placed the controller at location $5$ and the MAV flied in a square trajectory due to the limited area. As shown in Fig.~\ref{fig:nlosflight}, the average position error over the trajectory is $52.56$ cm and the average orientation error is $6.64\degree$. The accuracy is slightly worse than in the open field due to the multipath fading and the more aggressive motions around the corners of the square trajectory. 

Overall, the performance is better than GPS-based state estimators~\cite{farrell2008aided, chao2010autopilots} (meter-level accuracy), which are only applicable in outdoors. In addition, the position accuracy is also better than the state-of-the-art WiFi indoor localization systems~\cite{kotaru2015spotfi, xie2019md, vasisht2016decimeter}, which have to use an active WiFi radio and cannot work in long-range or through-wall settings. Note that Marvel is not designed to defeat CV-based approaches~\cite{qin2017vins, lin2018autonomous, zhu2017event} in accuracy. They are more accurate when environments are well-lighted and texture-rich. But Marvel complements them to support MAV navigation in vision-crippled scenarios. 

%% file: related.tex
State estimation for aerial vehicles has been a long studied problem in the robotics community. LiDAR and camera~\cite{qin2017vins, hess2016real, lin2018autonomous, zhu2017event, dube2017online, mur2015orb} are the representative sensors to enable state estimation. LiDAR is suitable for standard-size aerial vehicles due to its heavy weight and high cost. Although camera is more acceptable for MAVs due to its lightweight and high accuracy, it is limited in well-lighted and texture-rich environments, hindering its usage in vision-crippled scenarios, \eg, smoky buildings in firefighting operations.

To ease the limitation of existing solutions, RF-based state estimators have been proposed in that RF signals are highly resilient to visual limitations. Mueller~\et~\cite{mueller2015fusing} and Liu~\et~\cite{liu2017cooperative} take advantage of UWB-based ranges to enable state estimation. WINS~\cite{zhang2018wins} uses ubiquitous WiFi to estimate angle-of-arrivals (AoAs) upon an onboard antenna array for state estimation. Extensive studies of WiFi indoor localization systems~\cite{kotaru2015spotfi, xie2019md, vasisht2016decimeter} in the networking community also demonstrate its potential in state estimation. These proposals operate correctly where the amplitude and phase of RF signals are available but not in the presence of extremely weak signals that are far below the noise floor. 

Recently, communications with low-power signals in long-range or occlusive settings have been studied in~\cite{talla2017lora, peng2018plora, varshney2017lorea}. The signal characteristic and the processing method enable the localizability with such low-power signals. $\mu$locate~\cite{nandakumar20183d} is the first localization system that extracts the channel phases of low-power CSS signals drowned by noise to localize targets by range estimates. It operates correctly in semi-stationary scenarios but not in the presence of agile mobilities of MAVs. The fundamental difference in our context is that the backscattered CSS signals have Doppler frequency shifts. Moreover, $\mu$locate only addresses location and requires the floor plan of the work space to localize targets with a single access point (AP). In contrast, Marvel designs novel algorithms to 1) localize a MAV with its single controller without any prior knowledge of the work space; 2) track the MAV rotation via low-power backscattered CSS signals; 3) enable accurate state estimation by a backscatter-inertial super-accuracy algorithm with the aid of IMU. In addition, there are also works devoted to aided inertial sensing~\cite{liu2017cooperative, qin2017vins, lupton2012visual}, but they are much different from our work due to the fundamentally different sensing modality -- backscatter-based pose sensing -- formulating a new measurement model to fuse with IMU.

%% file: conclusion.tex
To our knowledge, Marvel represents the first RF backscatter-based MAV state estimation system that works in long range or through wall with low-power signals drowned by noise. It marks a new sensing modality that complements existing visual solutions in supporting MAV navigation. The system is powered by a backscatter-based pose sensing module that estimates pose features via backscattered CSS signals as well as a backscatter-inertial super-accuracy algorithm that leverages IMU for accurate state estimation. We implement Marvel on USRP and the DJI Matrice 100 platform with customized backscatter tags. The experimental results based on three flight trajectories in both outdoors and indoors show that Marvel holds the promise as a long-range/through-wall, lightweight and plug-and-play state estimation system for MAVs. In future, we plan to seamlessly combine visual sensing and low-power RF sensing to achieve a more robust state estimation system.